%% file: main.tex
\documentclass[10pt,twocolumn,letterpaper]{article}

\makeatletter
\g@addto@macro\normalsize{%
  \setlength{\abovedisplayskip}{4pt}%
  \setlength{\belowdisplayskip}{4pt}%
  \setlength{\abovedisplayshortskip}{2pt}%
  \setlength{\belowdisplayshortskip}{2pt}%
}
\makeatother

\usepackage{cvpr}              


\definecolor{cvprblue}{rgb}{0.21,0.49,0.74}
\usepackage[pagebackref,breaklinks,colorlinks,allcolors=cvprblue]{hyperref}

\def\paperID{1406} 
\def\confName{CVPR}
\def\confYear{2025}

\title{OmniPSD: Layered PSD Generation with Diffusion Transformer}

\author{
  Cheng Liu\textsuperscript{1, *} \quad
  Yiren Song\textsuperscript{1, *} \quad
  Haofan Wang\textsuperscript{2} \quad
  Mike Zheng Shou\textsuperscript{1 $\dagger$} \\
  \textsuperscript{1}National University of Singapore \quad
  \textsuperscript{2}Lovart AI
}

\begin{document}

\twocolumn[{%
\renewcommand\twocolumn[1][]{#1}%
\maketitle
\vspace{-5mm}
\centering
\includegraphics[width=\textwidth]{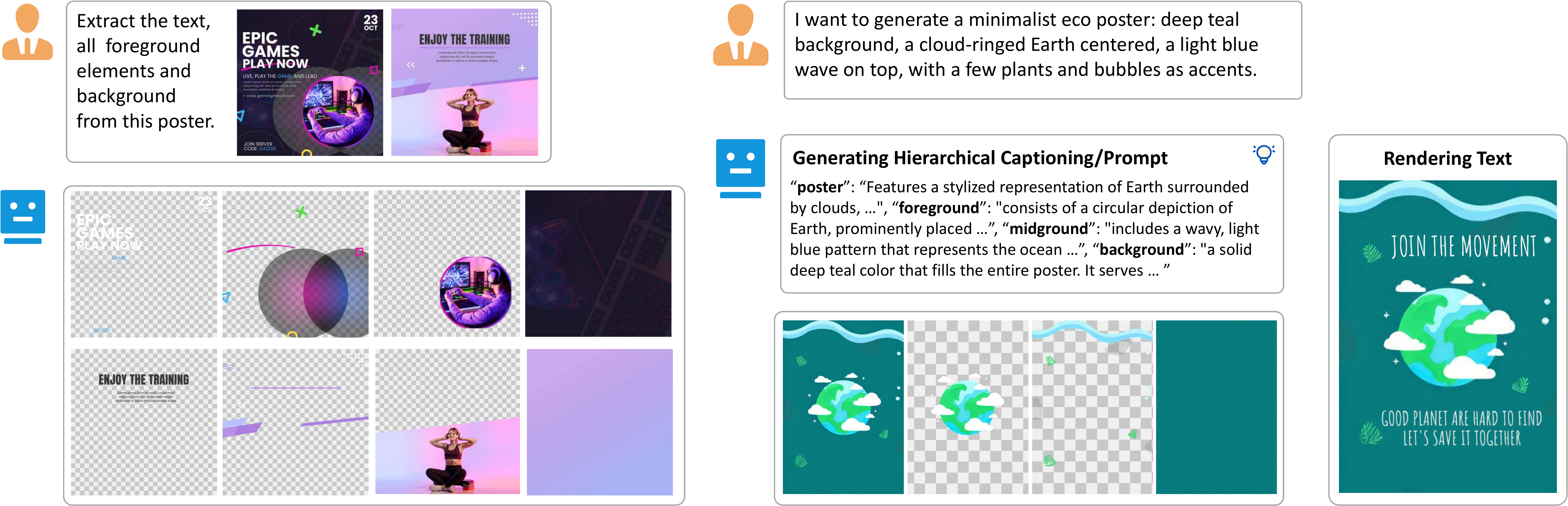}
\vspace{-5mm}
\captionof{figure}{
\textbf{OmniPSD} is a Diffusion-Transformer framework that generates layered PSD files with transparent alpha channels. Our system supports both \textit{Text-to-PSD} multi-layer synthesis and \textit{Image-to-PSD} reconstruction, producing editable layers that preserve structure, transparency, and semantic consistency.
}
\vspace{1.0em}
\label{fig:teaser}
}]

\begingroup
\renewcommand\thefootnote{}
\footnotetext{* Equal contribution.}
\footnotetext{$\dagger$ Corresponding author.}
\endgroup

\maketitle
\input{sec/0_abstract}

\input{sec/1_intro}

\input{sec/2_formatting}
\input{sec/3_finalcopy}

{
    \small
    \bibliographystyle{ieeenat_fullname}
    \bibliography{main}
}


\clearpage
\appendix
\input{sec/4_Appendix}

\end{document}

%% file: sec/0_abstract.tex
\begin{abstract}
Recent advances in diffusion models have greatly improved image generation and editing, yet generating or reconstructing layered PSD files with transparent alpha channels remains highly challenging.
We propose OmniPSD, a unified diffusion framework built upon the Flux ecosystem that enables both text-to-PSD generation and image-to-PSD decomposition through in-context learning.
For text-to-PSD generation, OmniPSD arranges multiple target layers spatially into a single canvas and learns their compositional relationships through spatial attention, producing semantically coherent and hierarchically structured layers.
For image-to-PSD decomposition, it performs iterative in-context editing—progressively extracting and erasing textual and foreground components—to reconstruct editable PSD layers from a single flattened image.
An RGBA-VAE is employed as an auxiliary representation module to preserve transparency without affecting structure learning.
Extensive experiments on our new RGBA-layered dataset demonstrate that OmniPSD achieves high-fidelity generation, structural consistency, and transparency awareness, offering a new paradigm for layered design generation and decomposition with diffusion transformers. Project page: \href{https://showlab.github.io/OmniPSD/}{https://showlab.github.io/OmniPSD/}.

\end{abstract}

%% file: sec/1_intro.tex

\section{Introduction}

Layered design formats such as Photoshop (PSD) files are essential in modern digital content creation, enabling structured editing, compositional reasoning, and flexible element-level manipulation.  
However, most generative models today can only output flat raster images \cite{ho2020denoising,rombach2022high,peebles2023scalable}, lacking the layer-wise structure and transparency information that are crucial for professional design workflows.  
To bridge this gap, we introduce \textbf{OmniPSD}, a unified diffusion-based framework that supports both \textit{text-to-PSD generation} and \textit{image-to-PSD decomposition} under a single architecture.  
It enables bidirectional transformation between textual or visual inputs and fully editable, multi-layer PSD graphics.

At the core of OmniPSD lies a pre-trained \textbf{RGBA-VAE}, designed to encode and decode transparent images into a latent space that preserves alpha-channel information \cite{layerdiffuse2023,pu2025art,psdiffusion2025}.  
This RGBA-VAE serves as a shared foundation across both sub-tasks.  
On top of it, we leverage the \textbf{Flux ecosystem}, which consists of two complementary diffusion transformer models:  
\textbf{Flux-dev} \cite{flux2024}, a text-to-image generator for creative synthesis, and \textbf{Flux-Kontext} \cite{batifol2025flux}, an image editing model for in-context refinement and reconstruction.  
By integrating these components, OmniPSD provides a unified, transparency-aware solution for both generation and decomposition.

\noindent\textbf{(1) Text-to-PSD Generation.}  
Given a textual description, OmniPSD generates a layered PSD representation directly from text.  
Instead of producing a single flat image, we spatially arrange multiple semantic layers (e.g., background, foreground, text, and effects) into a $2\times2$ grid, and generate them simultaneously through the \textbf{Flux-dev} backbone.  
Each generated layer is then decoded by the shared RGBA-VAE to recover transparency and alpha information, producing semantically coherent, editable, and compositional layers.

\noindent\textbf{(2) Image-to-PSD Decomposition.}  
For reverse-engineering real or synthetic posters into editable PSDs, OmniPSD extends the \textbf{Flux-Kontext} model by replacing its standard VAE with our pre-trained RGBA-VAE, enabling transparency-aware reasoning in image editing.  
The decomposition process is iterative: we first extract text layers through in-context editing, then erase them to reconstruct the clean background, and finally segment and refine multiple foreground layers.  
All decomposed outputs are in RGBA format, ensuring accurate transparent boundaries and realistic compositional relationships.

By unifying \textit{generation} and \textit{decomposition} within a single diffusion-transformer architecture, OmniPSD demonstrates that both creative synthesis and structural reconstruction can be achieved under a transparency-aware, in-context learning framework.

Our main contributions are summarized as follows:

\begin{itemize}
\item 
We present \textbf{OmniPSD}, a unified diffusion-based framework that supports both text-to-PSD generation and image-to-PSD decomposition within the same architecture, bridging creative generation and analytical reconstruction.

\item 
We pre-train a transparency-preserving \textbf{RGBA-VAE} and integrate it with Flux-dev and Flux-Kontext through in-context learning, achieving high-fidelity image generation and reconstruction with accurate alpha-channel representation.

\item  
We construct a large-scale dataset with detailed RGBA layer annotations and establish a new benchmark for editable PSD generation and decomposition.  
Extensive experiments demonstrate the effectiveness and superiority of our proposed approach.
\end{itemize}

\section{Related Works}

\begin{figure*}[t]
    \centering
    \includegraphics[width=\linewidth]{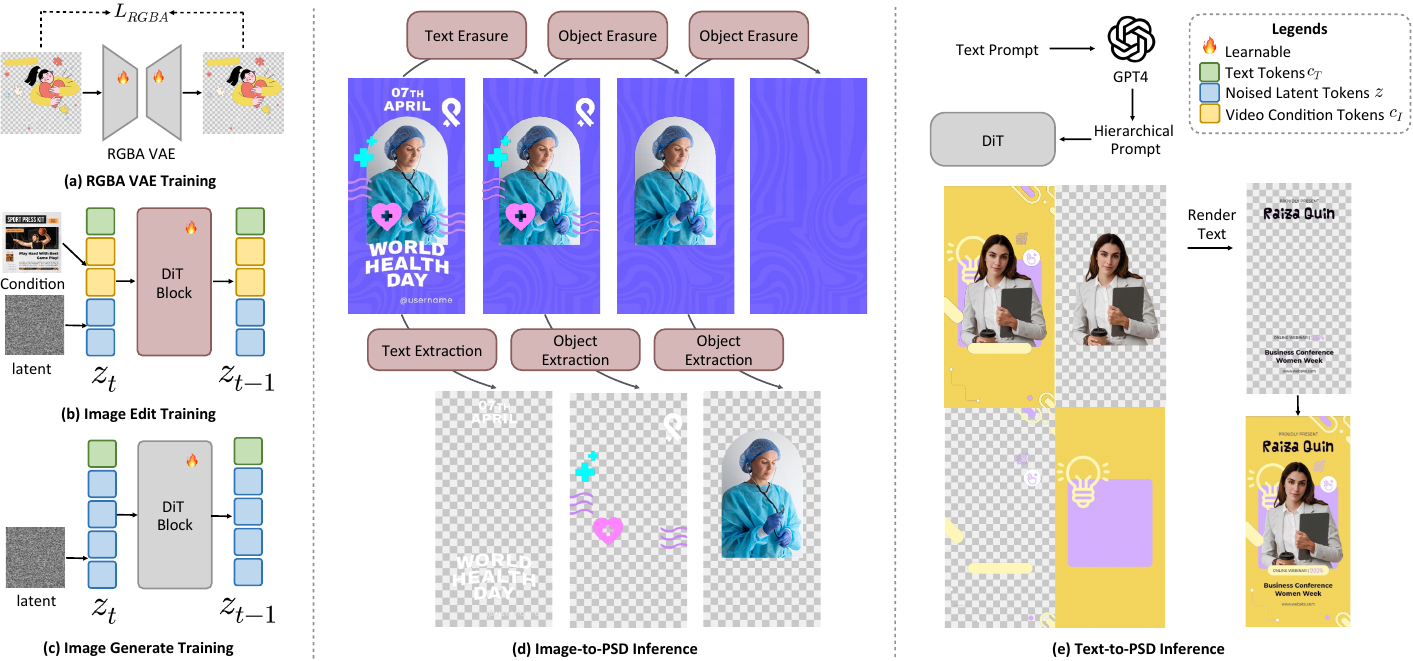}
    \caption{OmniPSD overview. A unified Diffusion-Transformer with a shared RGBA-VAE enables both text-to-PSD layered generation (left) and image-to-PSD decomposition (right). Text-to-PSD leverages spatial in-context learning with hierarchical captions, while Image-to-PSD performs iterative flow-guided foreground extraction and background restoration. Our method produces fully editable PSD layers with transparent alpha channels.}
    \label{fig:method}
\end{figure*}

\subsection{Diffusion Models}

Diffusion probabilistic models have rapidly become the dominant paradigm for high-fidelity image synthesis, largely replacing GANs due to their stable training, strong mode coverage, and ability to model complex data distributions via reversed noising processes~\cite{song2020denoising,ho2020denoising,goodfellow2020generative}. They now underpin a broad range of visual tasks, including text-to-image generation \cite{zhang2024ssr, zhang2024stable, zhang2025stable, song2025makeanything}, image editing \cite{song2025omniconsistency, jiang2025personalized, gong2025relationadapter, huang2025arteditor}, and video synthesis~\cite{rombach2022high,ho2022video,blattmann2023stable, song2024processpainter, ma2024followpose, ma2024followyouremoji, ma2025followyourclick, ma2025followyourmotion}. To better support design and editing applications, subsequent work augments diffusion models with grounded conditions and spatial controls---for example, grounded text-to-image generation, conditional control branches, image-prompt adapters, cross-attention-based prompt editing, self-guided sampling, inpainting modules, as well as instance-level and multi-subject layout control~\cite{li2023gligen,zhang2023controlnet,ye2023ip,hertz2022prompt,epstein2023diffusionself,yu2023inpaintanything,cao2023masactrl,wang2024instancediffusion,wang2025msdiffusion}---thereby improving layout consistency and local editability. Early work predominantly relied on U-Net-based denoisers in pixel or latent space, as popularized by latent diffusion models such as Stable Diffusion and SDXL~\cite{rombach2022high,podell2023sdxl}. More recently, Transformer-based denoisers have become the de facto backbone, with Diffusion Transformers (DiT) driving models like Stable Diffusion 3, FLUX, HunyuanDiT, and PixArt, leveraging global attention and scalability to improve visual fidelity and prompt alignment~\cite{peebles2023scalable,esser2024sd3,flux2024,hunyuan2024,chen2023pixart}. In parallel, flow-matching and ODE-based formulations recast diffusion as learning continuous deterministic flows between distributions, enabling more efficient sampling and deterministic trajectories~\cite{lipman2022flow,liu2022flow}.

\subsection{Layer-wise Image Generation}

Layered image representations are fundamental to graphics and design, as they enable element-wise editing, compositional reasoning, and asset reuse \cite{lee2020hybrid}. Early work mainly decomposes a single image into depth layers, alpha mattes, or semantic regions under simplified foreground–background assumptions \cite{xu2017deep,sengupta2020background,li2022bridging,chen2022ppmatting}, which helps matting and segmentation but falls short of the rich, editable layer structures used in professional tools. With diffusion models, newer methods explicitly target layered generation. Some methods still rely on post-hoc detection, segmentation, and matting from a flat RGB output, or adopt a two-stage ``generate-then-decompose'' pipeline, where a composite RGB image is first synthesized and then separated into foreground/background layers or RGBA instances~\cite{liu2022learning,zhang2023text2layer,kang2025layeringdiff,fontanella2024rgba,wang2025diffdecompose}. Such designs often accumulate errors between stages and offer limited control over global layout and inter-layer relationships.

More recent approaches generate multi-layer content directly in a diffusion framework. LayerDiff, LayerDiffuse, and LayerFusion explore layer-collaborative or multi-branch architectures to jointly synthesize background and multiple foreground RGBA layers while modeling occlusion relationships~\cite{yang2023layerdiff,zhang2024layerdiffuse,dalva2024layerfusion}. ART and LayerTracer further introduce region-based transformers and vector-graphic decoders for variable multi-layer layouts and object-level controllability~\cite{pu2025art,song2025layertracer}. In parallel, multi-layer datasets such as MuLAn provide high-quality RGBA annotations and occlusion labels to support controllable multi-layer generation and editing~\cite{tudosiu2024mulan}. Recent works like PSDiffusion explicitly harmonize layout and appearance across foreground and background layers~\cite{psdiffusion2025}, and our method follows this line while additionally targeting PSD-style layer structures and workflows tailored for poster and graphic design.

Orthogonal to transparent-layer modeling, another line of work in automatic graphic design and poster generation emphasizes layout- and template-level generation. COLE and OpenCOLE propose hierarchical pipelines that decompose graphic design into planning, layer-wise rendering, and iterative editing~\cite{jia2023cole,inoue2024opencole}. Graphist formulates hierarchical layout generation for multi-layer posters with a large multimodal model that outputs structured JSON layouts for design elements~\cite{cheng2024graphist}. Visual Layout Composer introduces an image–vector dual diffusion model that jointly generates raster backgrounds and vector elements for design layouts~\cite{shabani2024vlc}. MarkupDM and Desigen treat graphic documents as multimodal markup or controllable design templates, enabling completion and controllable template generation from partial specifications~\cite{kikuchi2025markupdm,weng2024desigen}. PosterLLaVa further leverages multimodal large language models to generate poster layouts and editable SVG designs from natural-language instructions~\cite{yang2024posterllava}. These systems focus on high-level layout synthesis but typically output flattened renders or coarse vector structures, whereas our approach targets PSD-style RGBA layers with explicit alpha channels, making the resulting assets directly editable and composable in professional design tools.

\subsection{RGBA Image Generation}

Generating transparent or layered RGBA content is crucial for compositing and design, yet has long been underexplored compared to standard RGB image synthesis. Traditional workflows typically rely on first generating opaque RGB images and then applying separate matting, segmentation, or alpha-estimation networks \cite{xu2017deep,sengupta2020background,li2022bridging,chen2022ppmatting,hu2024diffumatting,li2024mattinganything}, which often leads to inconsistent boundaries, halo artifacts, and limited control over transparency. Recent diffusion-based methods begin to treat transparency as a first-class signal. One representative line augments latent diffusion models with “latent transparency”, learning an additional latent offset that encodes alpha information while largely preserving the original RGB latent manifold, so that existing text-to-image backbones can natively produce transparent sprites or multiple transparent layers without retraining from scratch \cite{zhang2024layerdiffuse}. Building on this idea, RGBA-aware generators produce isolated transparent instances or sticker-like assets that can be flexibly composed for graphic design and poster layouts~\cite{fontanella2024rgba,quattrini2024alfie}. 

Complementary work focuses on the representation side, proposing unified RGBA autoencoders that extend pretrained RGB VAEs with dedicated alpha channels and introducing benchmarks that adapt standard RGB metrics to four-channel images via alpha compositing, thereby standardizing evaluation for RGBA reconstruction and generation \cite{alphavae2025}. Building on these ideas, multi-layer generation systems increasingly adopt autoencoders that jointly encode and decode stacked RGBA layers and couple them with diffusion transformers that explicitly model transparency and inter-layer effects \cite{layerdiffuse2023,dalva2024layerfusion,pu2025art,psdiffusion2025, chen2025transanimate, wang2025diffdecompose,yang2024layerdecomp}, often trained or evaluated on matting-centric multi-layer datasets such as MAGICK and MuLAn~\cite{burgert2024magick,tudosiu2024mulan}, yielding more accurate alpha boundaries, coherent occlusions, and realistic soft shadows in complex layered scenes.

%% file: sec/2_formatting.tex
\section{Method}

In this section, we first introduce  the unified OmniPSD architecture in Section~\ref{sec:overall}.
Next, Section~\ref{sec:rgba-vae} presents the RGBA-VAE module, which enables alpha-aware latent representation shared across both pathways.
Then, Section~\ref{sec:image2psd} discusses the Image-to-PSD process based on iterative in-context editing and structural decomposition.
After that, Section~\ref{sec:text2psd} describes the Text-to-PSD process, where layered compositions are generated via spatial in-context learning and cross-layer attention.
Finally, Section~\ref{sec:dataset} introduces the Layered Poster Dataset.

\subsection{Overall Architecture}
\label{sec:overall}

We propose OmniPSD, a unified diffusion-based framework designed to reconstruct and generate layered PSD structures from either raster images or textual prompts.
The framework is built upon the Flux model family \cite{flux2024, batifol2025flux}, combining Flux-Dev for text-to-image generation and Flux-Kontext for image editing within an in-context learning paradigm.
At its core, a shared RGBA-VAE provides an alpha-aware latent space, enabling consistent representation of transparency and compositional hierarchy across both generation and decomposition tasks.

Specifically, the Image-to-PSD branch iteratively decomposes a given poster into text, foreground, and background layers through LoRA-based editing under the Flux-Kontext backbone, ensuring accurate structural separation with preserved alpha channels.
In contrast, the Text-to-PSD branch arranges layers spatially within a single generation canvas, where the model learns inter-layer relations via spatial attention under the Flux-DEV backbone.
Together, these two pathways form a cohesive framework capable of bidirectional conversion between design images and editable PSD layers, supported by our large-scale Layered Poster Dataset for training and evaluation.

\begin{figure}[t]
    \centering
    \includegraphics[width=\linewidth]{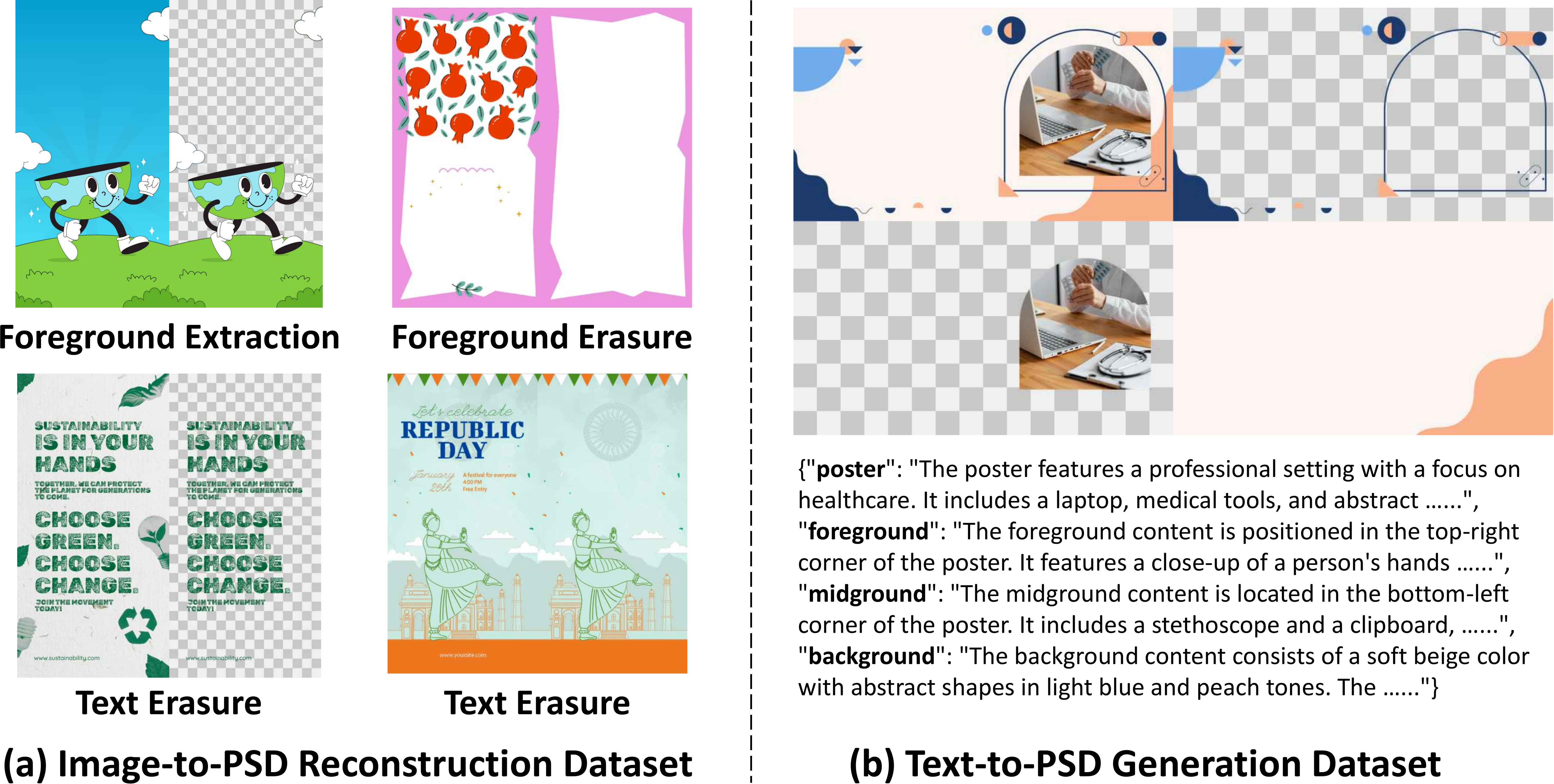}
    \caption{OmniPSD’s layered dataset. Image-to-PSD is trained on paired samples, while Text-to-PSD uses a $2\times2$ grid that presents the full poster and its decomposed layers for in-context learning.}
    \label{fig:Datasets}
    \vspace{-0.5cm}
\end{figure}

\subsection{RGBA-VAE}
\label{sec:rgba-vae}

To accurately represent transparency and compositional relationships in layered design elements, we adopt and extend the AlphaVAE~\cite{alphavae2025}, a unified variational autoencoder for RGBA image modeling.  
While AlphaVAE provides a strong foundation for alpha-aware reconstruction, its pretraining on limited natural transparency data causes severe degradation when applied to design scenarios such as semi-transparent text, shadow overlays, and soft blending effects.  
To address this, we retrain the model on our curated dataset of real-world design samples, enabling stable reconstruction of both alpha and color layers.  
We refer to this retrained version as RGBA-VAE.

Following the formulation in the original AlphaVAE paper, our training objective jointly optimizes pixel fidelity, patch-level consistency, perceptual alignment, and latent regularization as:
\begingroup
\setlength{\abovedisplayskip}{4pt}
\setlength{\belowdisplayskip}{4pt}
\begin{equation}
\begin{aligned}
\mathcal{L} =\;
& \lambda_{\text{pix}}\,
\mathbb{E}\!\left[\|\hat{I} - I\|_1\right]
+ \lambda_{\text{patch}}\,
\mathbb{E}\!\left[\|\phi(\hat{I}) - \phi(I)\|_1\right] \\
& + \lambda_{\text{perc}}\, 
\mathbb{E}\!\left[\|\psi(\hat{I}) - \psi(I)\|_2^2\right] \\
& + \lambda_{\text{KL}}\,
\Big(
\mathrm{KL}(q(z_{RGB}|\cdot)\|p)
+ \mathrm{KL}(q(z_{A}|\cdot)\|p)
\Big),
\end{aligned}
\end{equation}
\endgroup

where \(I\) and \(\hat{I}\) denote the ground-truth and reconstructed images, respectively.  
\(\phi(\cdot)\) represents a patch-level feature extractor enforcing local structure consistency,  
and \(\psi(\cdot)\) denotes a perceptual encoder (e.g., VGG) that maintains semantic fidelity.  
\(z_{RGB}\) and \(z_{A}\) correspond to the latent variables for color and alpha channels, and \(p\) is the Gaussian prior.  
The coefficients \(\lambda_{\text{pix}}\), \(\lambda_{\text{patch}}\), \(\lambda_{\text{perc}}\), and \(\lambda_{\text{KL}}\) balance pixel accuracy, local consistency, perceptual alignment, and latent regularization, respectively.  

This retraining procedure effectively bridges the gap between natural transparency modeling and design-layered imagery.  
The resulting RGBA-VAE thus provides a shared latent space for both our text-to-PSD and image-to-PSD modules, enabling high-fidelity, alpha-preserving decomposition and generation.

\subsection{Image-to-PSD Reconstruction}
\label{sec:image2psd}

We formulate the Image-to-PSD reconstruction task as a multi-step, iterative image-editing process, analogous to how professional designers manually decompose visual elements into layers in Photoshop. Instead of predicting all layers in a single pass, we progressively extract text and foreground objects, while recovering occluded background content. Each step outputs an RGBA PNG layer with accurate transparency. This iterative design ensures pixel-level fidelity, precise alpha recovery, and structural composability for final PSD reconstruction.

Concretely, we train two expert models: one specialized for foreground extraction and another for foreground removal and background restoration. After each extraction, the background-restoration model reconstructs clean background content, enabling the system to reveal deeper visual layers over iterations. Through this alternating “extract-foreground → erase-foreground” process, a flattened input image is gradually decomposed into a stack of text, foreground, and background layers suitable for PSD editing. This pipeline is built on the \textbf{Flux Kontext} diffusion backbone with task-specific LoRA adapters. The decomposition process is formulated as a conditional flow-matching problem, where the flattened image is treated as a conditioning input and the model learns a deterministic flow field that maps noisy latent states toward their target decomposed layer representations.

\noindent\textbf{Formulation.}  
Let $\mathbf{I}_0 \in \mathbb{R}^{H \times W \times 4}$ denote the flattened input poster image, and $\mathbf{y} \in \{\text{foreground}, \text{background}\}$ denote the target layer type.  
We define latent variables $\mathbf{z}_0 = E_{\alpha}(\mathbf{I}_0)$ and $\mathbf{z}_1 = E_{\alpha}(\mathbf{I}_y)$, where $E_{\alpha}$ is the RGBA-VAE encoder.  
Flux models the continuous transformation between $\mathbf{z}_0$ and $\mathbf{z}_1$ as a flow field $\mathbf{v}_\theta(\mathbf{z}_t, t \mid \mathbf{z}_0)$ governed by an ODE:
\begin{equation}
\frac{d\mathbf{z}_t}{dt} = \mathbf{v}_\theta(\mathbf{z}_t, t \mid \mathbf{z}_0), \quad t \in [0, 1],
\end{equation}
where $\mathbf{z}_t = (1-t)\mathbf{z}_0 + t\mathbf{z}_1$ represents intermediate latent states.

The training objective follows the standard \textbf{Flow Matching Loss}~\cite{lipman2022flow, flux2024}:
\begin{equation}
\mathcal{L}_{\text{flow}} = 
\mathbb{E}_{t \sim \mathcal{U}(0,1), (\mathbf{z}_0, \mathbf{z}_1)} 
\left\|
\mathbf{v}_\theta(\mathbf{z}_t, t \mid \mathbf{z}_0)
-
(\mathbf{z}_1 - \mathbf{z}_0)
\right\|_2^2,
\end{equation}
which enforces the learned flow field to align with the true displacement between input and target latents.  
This formulation avoids stochastic noise injection, leading to faster convergence and deterministic inference.

\noindent\textbf{Foreground Extraction Model.}  
Given $\mathbf{I}_0$, the model detects salient regions and generates RGBA layers for each foreground instance.  
Each LoRA adapter is trained on triplets $(\mathbf{I}_0, \mathbf{m}, \mathbf{I}_{\text{fg}})$, where $\mathbf{m}$ denotes a binary or bounding-box mask, and $\mathbf{I}_{\text{fg}}$ is the corresponding RGBA foreground target.  
Both conditional and target images are encoded into latent sequences:
\begin{equation}
\mathbf{z}_{\text{cond}} = E_{\alpha}(\mathbf{I}_0), \quad
\mathbf{z}_{\text{target}} = E_{\alpha}(\mathbf{I}_{\text{fg}}),
\end{equation}
then concatenated into a unified token sequence:
\begin{equation}
\mathbf{Z} = [\mathbf{z}_{\text{cond}}; \mathbf{z}_{\text{target}}].
\end{equation}

The transformer backbone applies \textbf{Multi-Modal Attention (MMA)} \cite{mma} with bidirectional context:
\begin{equation}
\mathbf{Z}' = \text{MMA}(\mathbf{Z}) = 
\text{Softmax}\!\left(\frac{QK^\top}{\sqrt{d}}\right)V,
\end{equation}

capturing pixel-level and semantic correlations between input and decomposed regions.  

\noindent \textbf{Foreground Erasure Model.}  
After extraction, we employ an erasure module trained to reconstruct occlusion-free backgrounds $\mathbf{I}_{\text{bg}}$ given the same condition $\mathbf{I}_0$ and mask $\mathbf{m}$.  
At each iteration $k$, the model removes the current foreground, restores the occluded background $\mathbf{I}_{\text{bg}}^{(k)}$, and stores the removed content $\mathbf{I}_{\text{fg}}^{(k)}$ as an independent RGBA layer:
\begin{equation}
\{\mathbf{I}_{\text{fg}}^{(1)}, \ldots, \mathbf{I}_{\text{fg}}^{(K)}, \mathbf{I}_{\text{bg}}\} \rightarrow \text{PSD Stack}.
\end{equation}
All LoRA modules share the same latent flow space of Flux Kontext, ensuring modular composability across text removal, object extraction, and background inpainting subtasks.

%

\noindent\textbf{Editable Text Layer Recovery.}
To transform rasterized text regions into editable design layers, we reconstruct vector-text through a unified OCR--font-recovery--rendering pipeline. We detect and recognize textual content from pixel-level inputs using a transformer-based OCR module, implemented via the open-source PaddleOCR toolkit \cite{paddleocr}, which provides state-of-the-art scene and document-text recognition with multilingual and layout-aware support. The recognized text regions are then associated with the most plausible typeface from a curated font bank through semantic font embedding retrieval, achieved using the lightweight font classify system \cite{storia_fontclassify}, which enables efficient deep-learning-based font matching across large-scale font libraries. The recovered text content together with its inferred font attributes is subsequently re-rendered as resolution-independent vector layers, yielding editable PSD text objects that faithfully preserve the original typography and layout structure.

\subsection{Text-to-PSD Generation}
\label{sec:text2psd}

While Image-to-PSD is highly effective at decomposing an existing image into layered RGBA components, in many real scenarios no reference image is available. Instead, users may wish to generate a fully layered PSD file \textit{directly from textual descriptions}. To meet this need, we introduce the \textbf{Text-to-PSD model}, which leverages hierarchical textual prompts, cross-modal feature alignment, and an in-context layer reasoning mechanism.

\noindent\textbf{In-Context Layer Reasoning via a 2$\times$2 Grid.}
Our key idea is to enable different layers to ``see'' each other without modifying the backbone or introducing explicit cross-layer attention modules \cite{wan2024grid}. We arrange four images—the full poster $\mathbf{I}_{\text{full}}$, foreground $\mathbf{I}_{\text{fg}}$, middle-ground $\mathbf{I}_{\text{mid}}$, and background $\mathbf{I}_{\text{bg}}$—into a 2$\times$2 grid:
\[
\mathbf{G} = 
\begin{bmatrix}
\mathbf{I}_{\text{full}} & \mathbf{I}_{\text{fg}} \\
\mathbf{I}_{\text{mid}} & \mathbf{I}_{\text{bg}}
\end{bmatrix}.
\]

This grid serves as an in-context visual canvas, enabling the model's native spatial attention to implicitly learn layer relationships such as layout consistency, occlusion ordering, color harmony, and transparency boundaries. During inference, the model generates all PSD layers jointly in a single pass.


\noindent\textbf{Hierarchical Text Prompts.}
To provide structured semantic grounding, we annotate each sample with a JSON record that assigns a dedicated description to the full poster and each semantic layer, e.g.,
\texttt{\{"poster": "...", "foreground": "...", "midground": "...", "background": "..." \}}.
Here, \texttt{poster} captures the global scene, while the remaining fields describe the corresponding layers.

\noindent\textbf{Grid Spatial In-Context Learning.}
The $2\times2$ grid $\mathbf{G}$ is encoded by the RGBA-VAE and processed by the DiT backbone in a single forward pass.
Spatial self-attention over this grid lets layer tokens attend to the full-poster tokens, so the model learns cross-layer correspondences and compositional relationships without any extra cross-layer modules.

\noindent\textbf{Training Objective.}
We retain the standard flow-matching objective of the diffusion transformer and introduce no additional losses, allowing the model to learn layered semantics purely from the hierarchical prompts and the in-context $2\times2$ grid formulation.

\begin{figure*}[t]
    \centering
    \includegraphics[width=\linewidth]{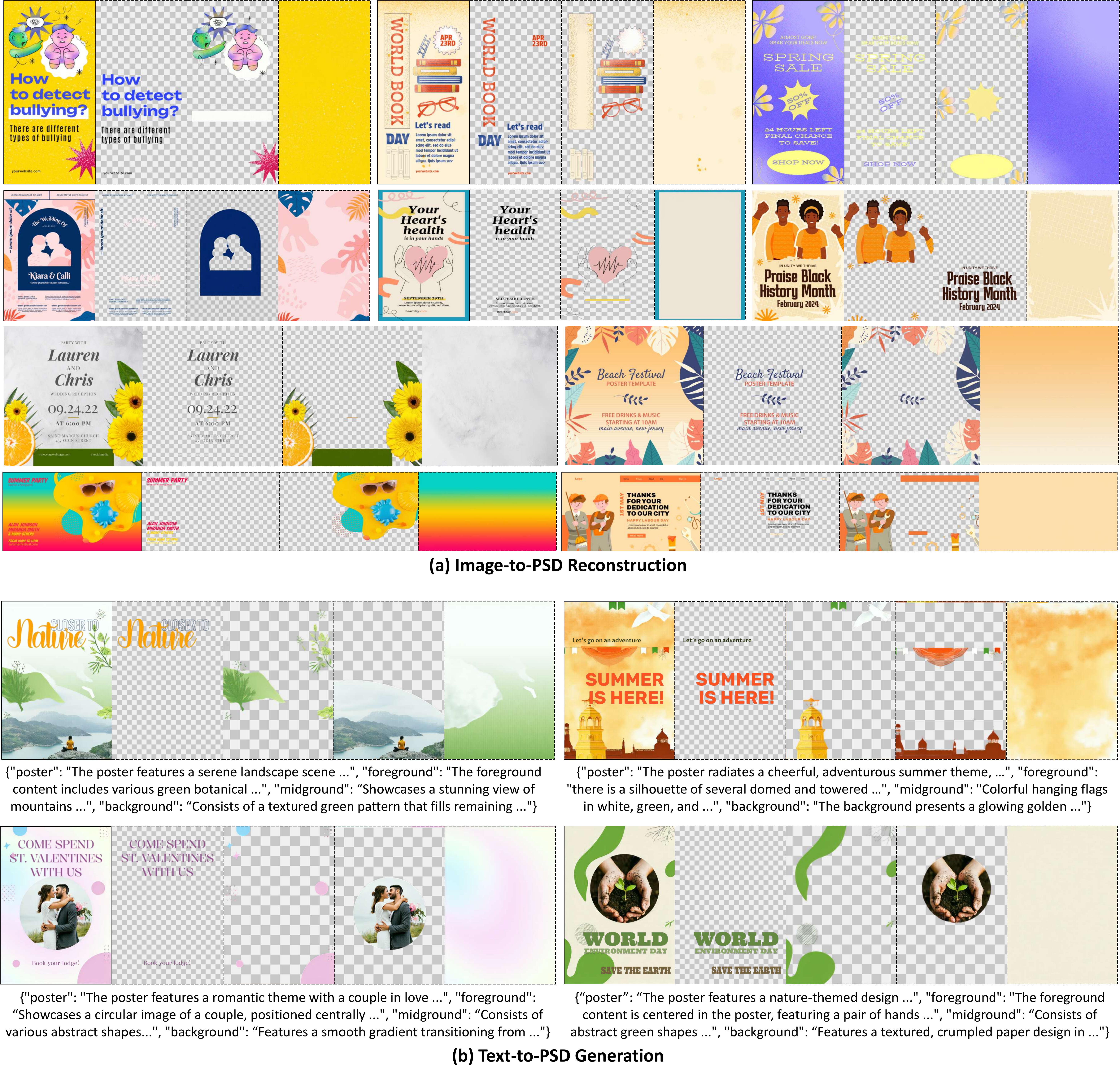}
    \vspace{-6mm}
    \caption{Generation results of OmniPSD. (a) \textbf{Image-to-PSD reconstruction} decomposes an input poster into editable text layers, multiple foreground layers, and a clean background layer. (b) \textbf{Text-to-PSD synthesis} uses hierarchical captions to generate background and foreground layers, followed by rendering the corresponding editable text layers.}
    \label{fig:Results}
\end{figure*}

\subsection{Dataset Construction}
\label{sec:dataset}

To support training and evaluation, we construct the Layered Poster Dataset, comprising over 200,000 real PSD files collected from online design repositories. These files are manually authored by professional designers and contain rich semantic groupings, font layers, shape groups, and effect overlays. We perform automated parsing to extract group-level and layer-level metadata, then apply post-filtering to retain only PSDs with valid RGBA structure. Each sample is annotated into structured groups—text, foreground, background—with each layer saved as an RGBA png and associated with editable metadata (e.g., bounding box, visibility, stacking order). 

To further support training across different subtasks, we organize the data with task-specific structures. For the Text-to-PSD generation task, we intentionally remove all text layers during dataset construction, since text should be rendered last rather than generated. This preserves authentic typography, font fidelity, and editability. The data is arranged in a four-panel grid: the top-left contains the full poster, while the remaining three panels provide semantic decomposition—top-right: foreground layer 1, bottom-left: foreground layer 2, and bottom-right: background layer. This format encourages the model to learn how text conditions map to layered design structures.

For the Image-to-PSD task, we adopt a triplet data strategy that mirrors the iterative layer editing process at inference time. Each triplet consists of (i) an input image, (ii) the extracted foreground content, and (iii) the corresponding background after foreground removal. This setup simulates the step-by-step editing workflow used in practical design software—first isolating editable regions, then erasing them from the scene—enabling the model to learn realistic PSD-style layer decomposition and reconstruction.

%% file: sec/3_finalcopy.tex
\section{Experiments}
\subsection{Experiment Details.}

\textbf{Experiment Details.}
During the Text-to-PSD training, we employed the Flux 1.0 dev model \cite{flux2024} built upon the pretrained Diffusion Transformer (DiT) architecture. The training resolution was set to 1024×1024 with a 2×2 grid layout. We adopted the LoRA fine-tuning strategy \cite{hu2022lora} with a LoRA rank of 128, a batch size of 8, a learning rate of 0.001, and 30,000 fine-tuning steps.

For the Image-to-PSD model training, we fine-tuned LoRA adapters on the Flux Kontext backbone \cite{batifol2025flux} at a resolution of 1024×1024. Specifically, we separately trained two types of modules—foreground extraction (for text and non-text elements) and foreground erasure (for text and non-text elements)—each for 30,000 steps.
For tasks that require transparency channels (e.g., Text-to-PSD, text extraction, and object extraction), we used the RGBA-VAE as the variational autoencoder.
For other tasks without transparency needs, we used the original VAE backbone.

\noindent \textbf{Baseline Methods.}  
    For the Text-to-PSD task, we benchmark against LayerDiffuse \cite{layerdiffuse2023} and GPT-Image-1 \cite{gpt}, the most relevant publicly available layered poster generation systems. For the Image-to-PSD task, to the best of our knowledge, this is the first work enabling editable PSD reconstruction from a single flattened image, and thus no prior method exists for direct comparison. Thus, we evaluate several commercial systems capable of producing RGBA layers \cite{gpt}, as well as a non-RGBA baseline \cite{batifol2025flux, nano_banana} where foregrounds are generated on a white canvas and transparency masks are derived using SAM2 segmentation \cite{ravi2024sam2}, representing a proxy solution without alpha-aware modeling.

\noindent \textbf{Metrics.}  
We evaluate OmniPSD using four metrics. FID \cite{fid} is computed on each generated layer and composite output to measure visual realism. For the Text-to-PSD task, we report layer-wise CLIP Score \cite{clip} to assess semantic alignment between each generated layer and its textual prompt. For the Image-to-PSD task, we compute reconstruction MSE by re-compositing predicted layers into a flattened image and measuring pixel error against the input. Together, these metrics capture realism, semantic fidelity, structural coherence, and reconstruction accuracy. To evaluate cross-layer structure and layout coherence, we employ GPT-4 \cite{gpt} as a vision-language judge, scoring spatial arrangement and design consistency. The detailed GPT-4 score metrics are provided in the supplementary materials \ref{sec:gpt4score}.

\noindent \textbf{Benchmarks.}  
For the Text-to-PSD task, we prepare a test set of 500 layer-aware prompts (two foreground, one background, and one global layout description), all derived from real PSD files to ensure realistic evaluation.  
For the Image-to-PSD task, we curate 500 real PSD files as the test set, which are flattened into single images for evaluating PSD reconstruction quality.

\noindent \textbf{User Study.}
We conducted a user study with 18 participants to evaluate the usability and perceptual quality of the layers generated by OmniPSD. The detailed study procedures and results are provided in the supplementary materials \ref{sec:userstudy}.

\begin{figure*}[t]
    \centering
    \includegraphics[width=\linewidth]{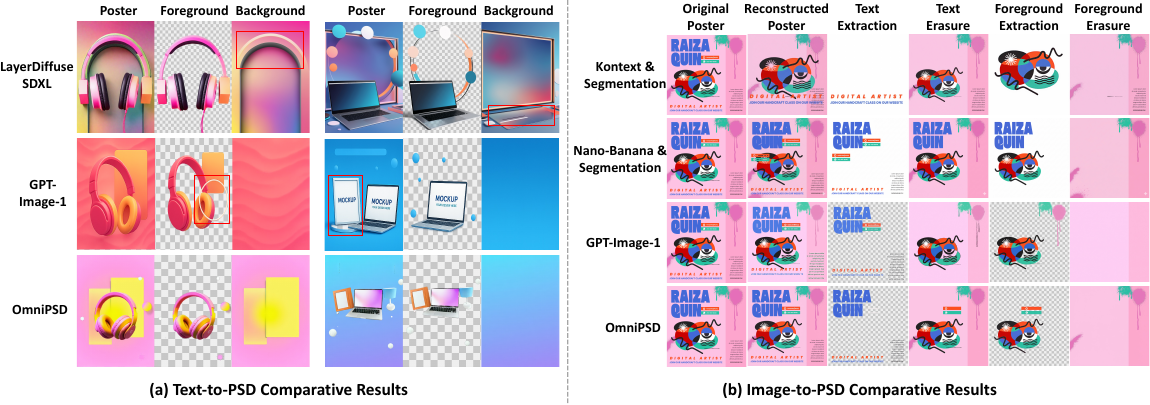}
    \caption{Compare with baselines on text-to-PSD and image-to-PSD. OmniPSD matches the visual quality of leading diffusion and vision-language models while uniquely supporting multi-layer PSD generation with transparent alpha channels. Compared to existing layered synthesis baselines, it achieves clearly superior visual fidelity and more coherent, logically structured layers.}
    \label{fig:Com}
\end{figure*}


\begin{table}[t]
\centering
\footnotesize
\setlength{\tabcolsep}{3pt}
\caption{Image-to-PSD generation results across methods. Lower is better for MSE; higher is better for PSNR, SSIM, CLIP-I (CLIP image score), and GPT-4-score. Bold numbers indicate the best performance for each metric.}
\vspace{-2mm}
\label{tab:com_img}
\begin{tabular}{lccccc}
\toprule
Method & MSE $\downarrow$ & PSNR $\uparrow$ & SSIM $\uparrow$ & CLIP-I $\uparrow$ & GPT-4-score $\uparrow$ \\
\midrule
Kontext \cite{batifol2025flux}  & 1.10e-1 & 9.59  & 0.653 & 0.692 & 0.64 \\
Nano-Banana \cite{nano_banana}  & 2.06e-2 & 16.9  & 0.816 & 0.916 & 0.86 \\
GPT-Image-1~\cite{gpt}      & 2.48e-2 & 16.1  & 0.761 & 0.837 & 0.84 \\
\textbf{OmniPSD (ours)}     & \textbf{1.14e-3} & \textbf{24.0} & \textbf{0.952} & \textbf{0.959} & \textbf{0.92} \\
\bottomrule
\end{tabular}
\end{table}

\begin{table}[t]
\centering
\footnotesize
\caption{Text-to-PSD generation results across methods. Lower is better for FID; higher is better for CLIP and GPT-4 scores. Bold numbers indicate the best performance for each metric.}
\vspace{-2mm}
\label{tab:com_txt}
\begin{tabular}{lccc}
\toprule
Method & FID $\downarrow$ & CLIP Score $\uparrow$ & GPT-4 Score $\uparrow$ \\
\midrule
LayerDiffuse SDXL~\cite{layerdiffuse2023} & 89.35  & 24.78  & 0.66  \\
GPT-Image-1~\cite{gpt}               & 53.21  & 35.59  & 0.84  \\
\textbf{OmniPSD (ours)}              & \textbf{30.43} & \textbf{37.64} & \textbf{0.90} \\
\bottomrule
\end{tabular}
\end{table} 

\begin{table}[t]
\centering
\footnotesize
\setlength{\tabcolsep}{5pt}
\caption{Image-to-PSD evaluation. Lower is better for FID and MSE; higher is better for PSNR and GPT-4 scores. We evaluate two sub-tasks---foreground extraction and foreground erasure---as well as the full reconstruction pipeline.}
\vspace{-2mm}
\label{tab:img_res}
\begin{tabular}{lcccc}
\toprule
Task & FID $\downarrow$ & MSE $\downarrow$ & PSNR $\uparrow$ & GPT-4 Score $\uparrow$ \\
\midrule
Text Extraction         & 11.42 & 1.34e-3 & 26.86 & 0.86 \\
Text Erasure            & 19.38 & 1.15e-3 & 26.37 & 0.94 \\
Foreground Extraction   & 33.35 & 2.26e-3 & 19.27 & 0.84 \\
Foreground Erasure      & 27.14 & 2.13e-3 & 29.41 & 0.92 \\
\midrule
Full Image-to-PSD     & 24.71  & 1.14e-3 & 23.98 & 0.90 \\
\bottomrule
\end{tabular}
\end{table}


\subsection{Comparison and Evaluation}

\noindent \textbf{Qualitative Evaluation.}
Figure \ref{fig:Results} and \ref{fig:Com} show the qualitative comparison.
For text-to-PSD, LayerDiffuser-SDXL produces plausible foregrounds and layouts but unstable, artifact-prone backgrounds, while GPT-Image-1, despite strong visual quality, often loses or alters background elements, harming global consistency. OmniPSD, by contrast, yields high-quality foreground and background layers with coherent overall posters. For image-to-PSD, baselines do not output true RGBA layers and thus cannot provide checkerboard visualizations. OmniPSD accurately performs text extraction, foreground extraction/removal, and background reconstruction, whereas other methods struggle to recover text and maintain consistency between extracted and erased regions, limiting their usability for PSD-style editing.

\noindent \textbf{Quantitative Evaluation.} This section presents quantitative analysis results. Table \ref{tab:com_img} and \ref{tab:com_txt} summarize the comparison results. Table \ref{tab:img_res} further reports the performance of each component in the image-to-PSD pipeline.
 Compared with strong baselines, OmniPSD achieves visual generation quality on par with state-of-the-art large diffusion and vision-language models. More importantly, our method uniquely supports multi-layer PSD generation with transparent alpha channels, a capability that existing approaches are far from achieving. Relative to prior layered synthesis systems, OmniPSD also demonstrates significant advantages in visual fidelity, semantic coherence, and logical layer structure, producing clean, editable layers that better reflect real design workflows.

\subsection{Ablation Study.}
 
\begin{figure}[ht]
    \centering
    \includegraphics[width=\linewidth]{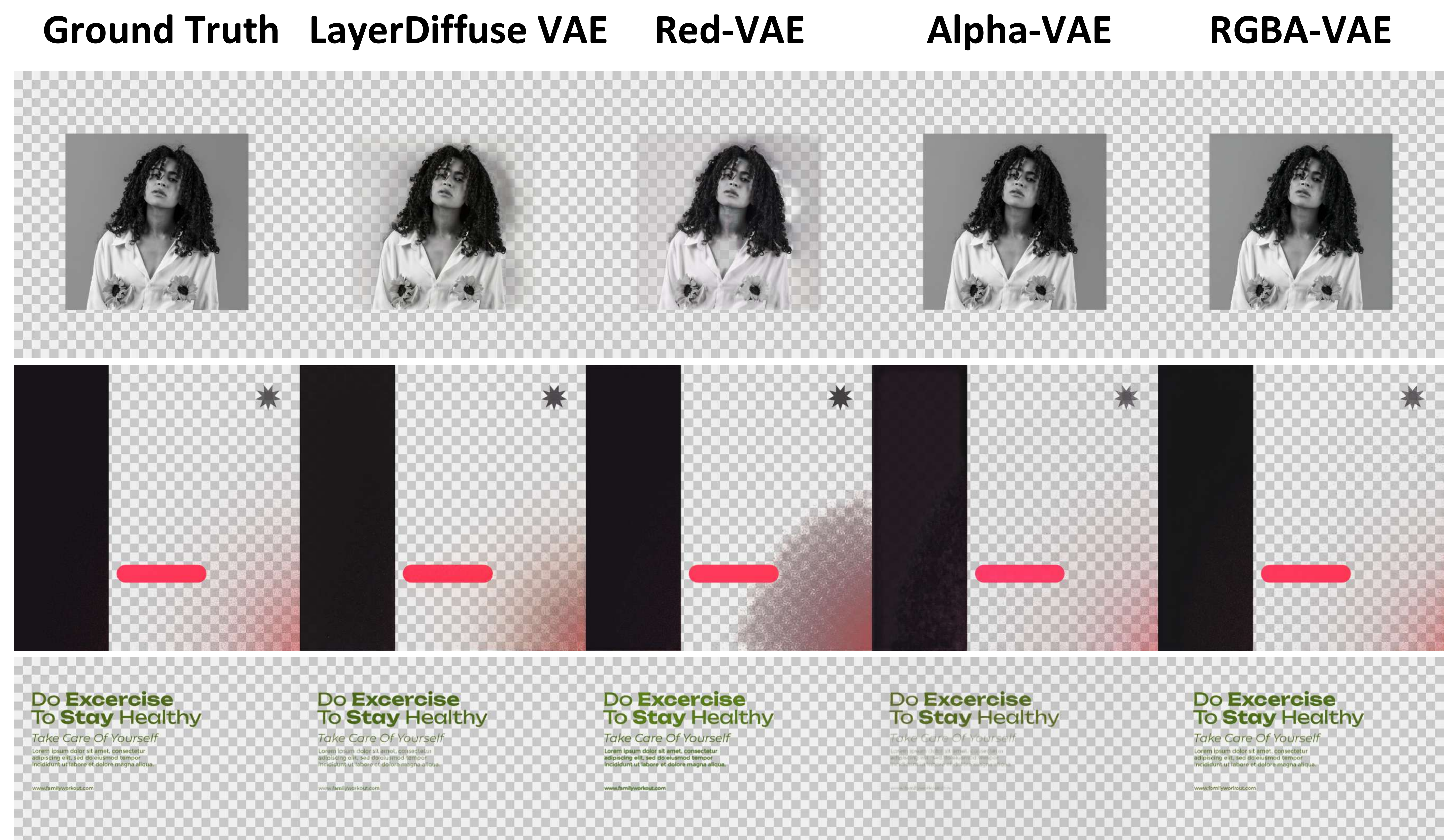}
    \caption{OmniPSD's RGBA-VAE. Compared to existing VAE methods which compatible with image alpha channels.}
    \label{fig:rgba}
\end{figure}

In this section, we present a detailed ablation study. We first compare our RGBA-VAE with other VAEs capable of encoding and decoding alpha channels. As shown in Table \ref{tab:rgba} and \ref{fig:rgba}, models trained primarily on natural images perform poorly in the design-poster setting, exhibiting inconsistent reconstruction, noticeable artifacts, and blurred text.
Table \ref{tab:Ablation} further highlights the importance of structured, layer-wise prompts in the text-to-PSD task: when using naive prompts, the generation quality degrades significantly.

\begin{table}[t]
\centering
\footnotesize
\caption{RGBA reconstruction results. Lower is better for MSE and LPIPS; higher is better for PSNR and SSIM. Bold numbers indicate the best performance for each metric.}
\vspace{-2mm}
\label{tab:rgba}
\begin{tabular}{lcccc}
\toprule
Method & MSE $\downarrow$ & PSNR $\uparrow$ & SSIM $\uparrow$ & LPIPS $\downarrow$ \\
\midrule
LayerDiffuse VAE~\cite{layerdiffuse2023} & 2.54e-1 & 8.06 & 0.289 & 0.473 \\
Red-VAE~\cite{layerdiffuseflux2025}      & 2.52e-1 & 8.53 & 0.300 & 0.451 \\
Alpha-VAE~\cite{alphavae2025}            & 4.15e-3    & 26.9 & 0.739 & 0.120 \\
\textbf{RGBA-VAE (ours)}                 & \textbf{9.82e-4} & \textbf{32.5} & \textbf{0.945} & \textbf{0.0348} \\
\bottomrule
\end{tabular}
\end{table}

\begin{table}[t]
\centering
\footnotesize
\caption{Ablation study results in  Text-to-PSD task. }
\vspace{-2mm}
\label{tab:Ablation}
\begin{tabular}{lccc}
\toprule
Method & FID $\downarrow$ & CLIP Score $\uparrow$ & GPT-4 Score $\uparrow$ \\
\midrule
w/o layer-specific prompt & 38.56  & 34.31  & 0.78  \\
\textbf{OmniPSD full} & \textbf{30.43} & \textbf{37.64} & \textbf{0.90} \\
\bottomrule
\end{tabular}
\end{table}

\section{Conclusion}

In this paper, we present OmniPSD, a unified framework for layered and transparency-aware PSD generation from a single raster image.
Built upon a Diffusion Transformer backbone, OmniPSD decomposes complex poster-style images into structured RGBA layers through an iterative, in-context editing process.
Our framework integrates an RGBA-VAE for alpha-preserving representation and multiple task-specific Kontext-LoRA modules for text, object, and background reconstruction.
We further construct a large-scale, professionally annotated layered dataset to support training and evaluation.
Extensive experiments demonstrate that OmniPSD achieves superior structural fidelity, transparency modeling, and semantic consistency, establishing a new paradigm for design-aware image decomposition and editable PSD reconstruction.

%% file: sec/4_Appendix.tex
\section*{Supplementary}

In the supplementary material, we provide additional details on the GPT-4-based automatic evaluation protocol, describe the design and results of our user study in both text-to-PSD and image-to-PSD settings, present more qualitative examples of OmniPSD’s layered poster generation and reconstruction, and showcase the interactive user interface and typical editing workflows supported by our system.

\section{GPT-4 Evaluation}
\label{sec:gpt4score}

In this section, we provide additional details about the automatic GPT-4-based evaluation protocol used in our experiments. We describe how candidate layered posters from different methods are jointly presented to GPT-4, the discrete 1--5 scoring rubric, the JSON output format, and how we aggregate and normalize these scores to obtain the final quantitative metric reported in the main paper.

\noindent\textbf{Implementation details of the GPT-4 evaluation.}
We adopt GPT-4 as an automatic visual judge to assess the quality of layered posters produced by different methods. For each input (either a text description or an image), we collect all candidate outputs from the compared methods and submit them \emph{together} in a single query. GPT-4 then gives each method an independent score, which allows a fair, side-by-side comparison under exactly the same context.

The assistant evaluates the layered poster results of different methods according to a 1--5 scale:
\begin{itemize}
    \item 1 = very poor (severely unreasonable, chaotic structure, and strong visual inconsistency),
    \item 2 = poor,
    \item 3 = fair / acceptable but with clear flaws,
    \item 4 = good,
    \item 5 = very good (clear structure, reasonable layer relationships, and visually coherent as a whole).
\end{itemize}

Scores are output in JSON format, for example:
\begin{verbatim}
{
  "Method1": 4,
  "Method2": 5,
  "Method3": 4
}
\end{verbatim}

For each method in a given query, GPT-4 assigns one integer score within this range based on the overall visual quality of the layered poster, including the consistency between layers, the plausibility of occlusion and depth, and the readability and layout of the final composed poster. The reported GPT-4 score in the main paper is obtained by averaging these integer scores over all test cases and then linearly normalizing the result to $[0,1]$.

\medskip
\noindent\textbf{Example of task prompt and evaluation.}
\textbf{Prompt:} ``You will evaluate layered poster results produced by multiple methods under the same input. For each method, inspect the full composited poster, the text layer, the foreground layer(s), and the background layer, and assign a single integer score in \{1,2,3,4,5\} based on visual consistency between layers, plausibility of occlusion and depth, and readability and layout of the final composed poster.''

\noindent\textbf{Images:} [Upload the layered poster results]

\noindent\textbf{Evaluation:} The assistant scores each method from 1 to 5 and returns the result in JSON format.


\section{User Study}
\label{sec:userstudy}
This subsection gives additional information about the user study conducted to evaluate OmniPSD in both text-to-PSD and image-to-PSD settings. We describe the participant pool, the evaluation criteria, and the 5-point Likert rating protocol, and we summarize how subjective feedback from designers and students supports the quantitative improvements reported in the main paper.

\noindent\textbf{Text-to-PSD.}
In the text-to-PSD setting, participants compared OmniPSD with LayerDiffuse-SDXL and GPT-Image-1 on 50 text prompts. For each generated layered poster, they rated two criteria on a 5-point Likert scale:
(1) \textit{layering reasonableness} (whether foreground, background, and text are separated in a semantically meaningful way with plausible occlusion and depth), and
(2) \textit{overall preference} (the overall visual appeal and usability of the final composed poster, including readability and layout).
As summarized in Tab.~\ref{tab:user-text}, OmniPSD achieves the highest mean scores on both criteria, clearly outperforming the baselines.

\begin{table}[ht]
    \footnotesize
    \setlength{\tabcolsep}{3pt}
    \centering
    \caption{User study results for the text-to-PSD setting.}
    \vspace{-2mm}
    \label{tab:user-text}
    \begin{tabular}{lccc}
        \toprule
        Metric & LayerDiffuse-SDXL & GPT-Image-1 & OmniPSD  \\
        \midrule
        Layering reasonableness & 3.33 & 3.89 & 4.39 \\
        Overall preference      & 3.39 & 3.78 & 4.50 \\
        \bottomrule
    \end{tabular}
\end{table}

\noindent\textbf{Image-to-PSD.}
In the image-to-PSD setting, participants evaluated 50 poster images decomposed by OmniPSD and the same two baselines. For each decomposed result, they rated three criteria on a 5-point Likert scale:
(1) \textit{reconstruction consistency} (how well the recomposed poster from the predicted layers matches the original input image in content and structure),
(2) \textit{layering reasonableness} (whether the recovered layers form a clean and plausible decomposition with correct occlusion and depth), and
(3) \textit{overall preference} (the perceived quality and practical usability of the layered result as a design asset).
Tab.~\ref{tab:user-image} shows that OmniPSD again obtains the highest mean scores on all three criteria, with consistent gains over the baselines.

\begin{table}[ht]
    \footnotesize
    \setlength{\tabcolsep}{1.5pt}
    \centering
    \caption{User study results for the image-to-PSD setting.}
    \vspace{-2mm}
    \label{tab:user-image}
    \begin{tabular}{lcccc}
        \toprule
        Metric & Kontext & Nano-Banana & GPT-Image-1 & OmniPSD \\
        \midrule
        Reconstruction consistency & 3.05 & 4.06 & 4.11 & 4.56 \\
        Layering reasonableness    & 3.44 & 4.16 & 4.22 & 4.61 \\
        Overall preference         & 3.39 & 4.33 & 4.28 & 4.72 \\
        \bottomrule
    \end{tabular}
\end{table}

Across both settings, designers particularly praised OmniPSD for its ``clear layer separation'' and ``realistic transparency,'' which enables direct reuse in professional editing workflows. These results confirm that OmniPSD provides superior structural consistency and practical value for real-world design generation and reconstruction.


\section{More Results}
In this subsection, we present additional qualitative results of OmniPSD in both image-to-PSD reconstruction and text-to-PSD synthesis. These visual examples cover diverse layouts and contents, illustrating the clarity of the recovered layers, the realism of transparency and occlusion, and the overall visual quality of the final composed posters.

\begin{figure*}[t]
    \centering
    \includegraphics[width=\linewidth]{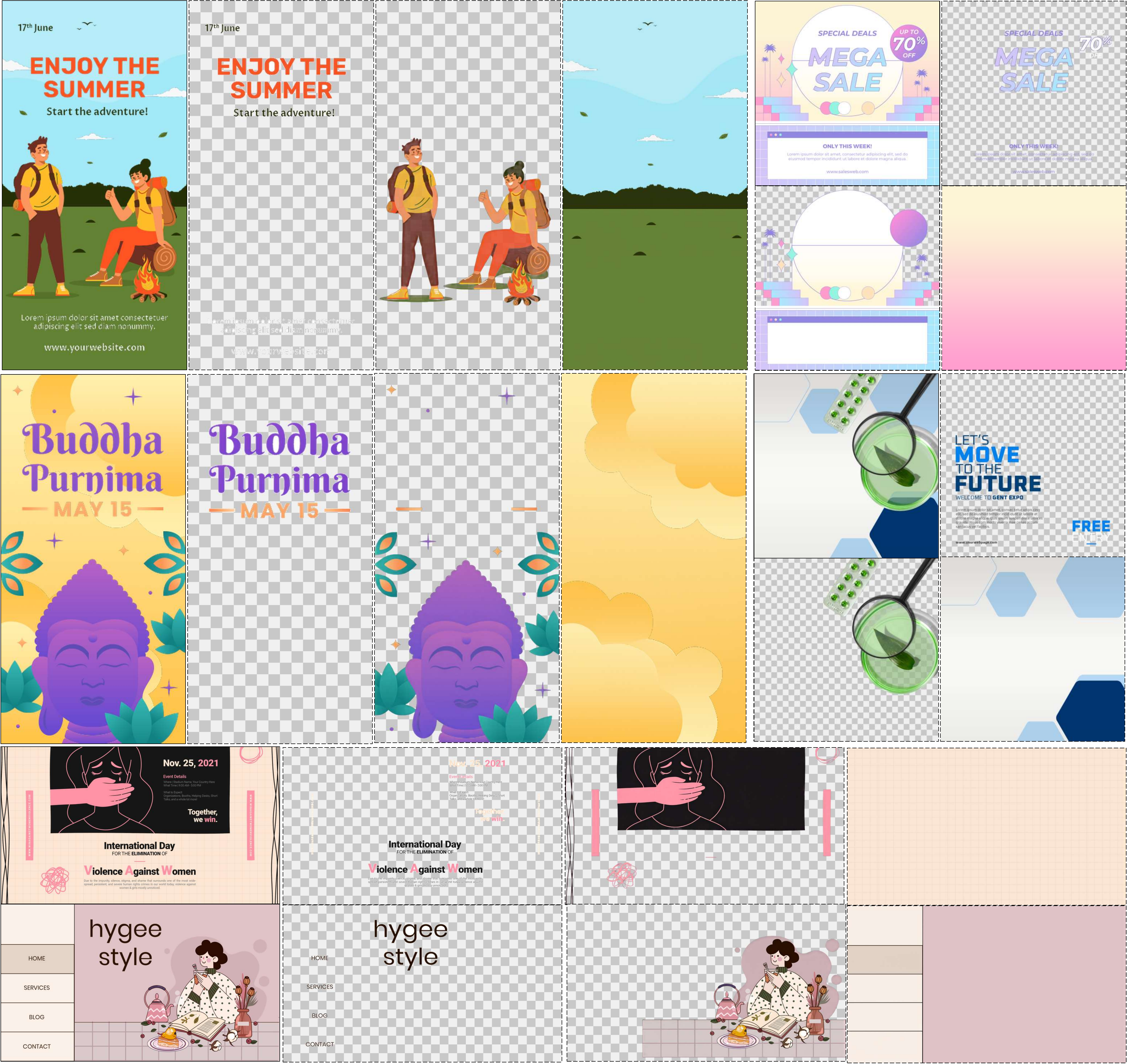}
    \caption{More generation results of OmniPSD image-to-PSD reconstruction.}
    \label{fig:more1}
\end{figure*}

\noindent\textbf{Image-to-PSD reconstruction.}
Figure~\ref{fig:more1} shows more examples where OmniPSD decomposes input poster images into layered PSD files and then recomposes them. The reconstructions exhibit high fidelity to the original designs while preserving clean layer boundaries that are convenient for downstream editing.

\begin{figure*}[t]
    \centering
    \includegraphics[width=0.85\linewidth]{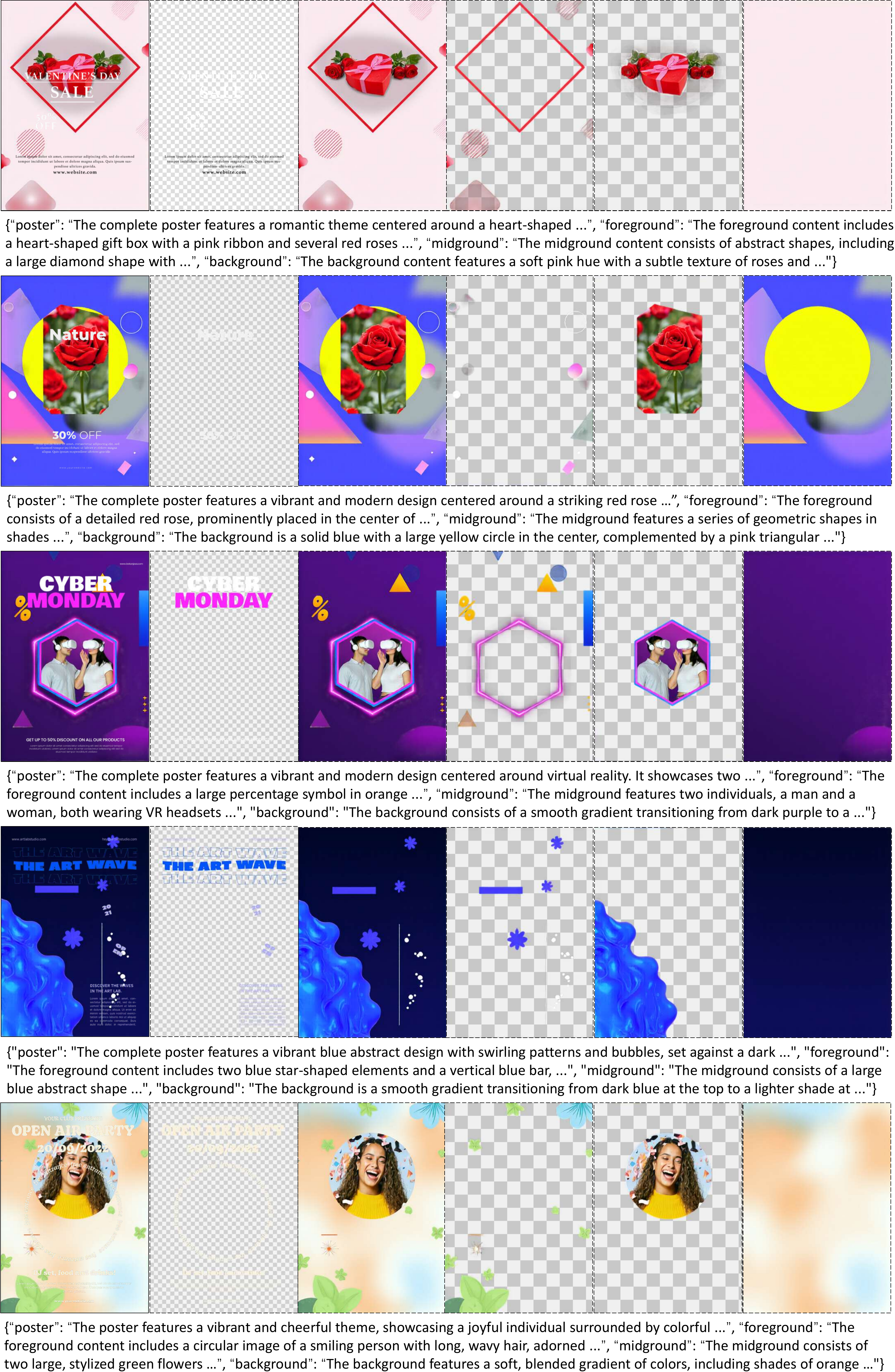}
    \caption{More generation results of OmniPSD text-to-PSD synthesis.}
    \label{fig:more2}
\end{figure*}

\noindent\textbf{Text-to-PSD synthesis.}
Figure~\ref{fig:more2} presents additional OmniPSD results in the text-to-PSD setting. Given only textual descriptions, our method synthesizes layered posters with coherent foreground elements, legible text, and visually consistent backgrounds, demonstrating its versatility as a generative design tool.

\section{User Interface}
In this subsection, we present the interactive user interface of OmniPSD and demonstrate typical editing workflows on a representative poster example. Starting from a user-uploaded image, OmniPSD automatically infers a layered representation that separates text, foreground objects, and background regions into editable components. Through intuitive point-and-click operations, users can modify textual content, remove or replace the background, and delete or adjust individual graphical elements while preserving the overall layout and visual coherence of the design. This interface illustrates how OmniPSD couples high-quality layer decomposition with practical, user-friendly tools for real-world poster editing and creation.

\begin{figure*}[t]
    \centering
    \includegraphics[width=\linewidth]{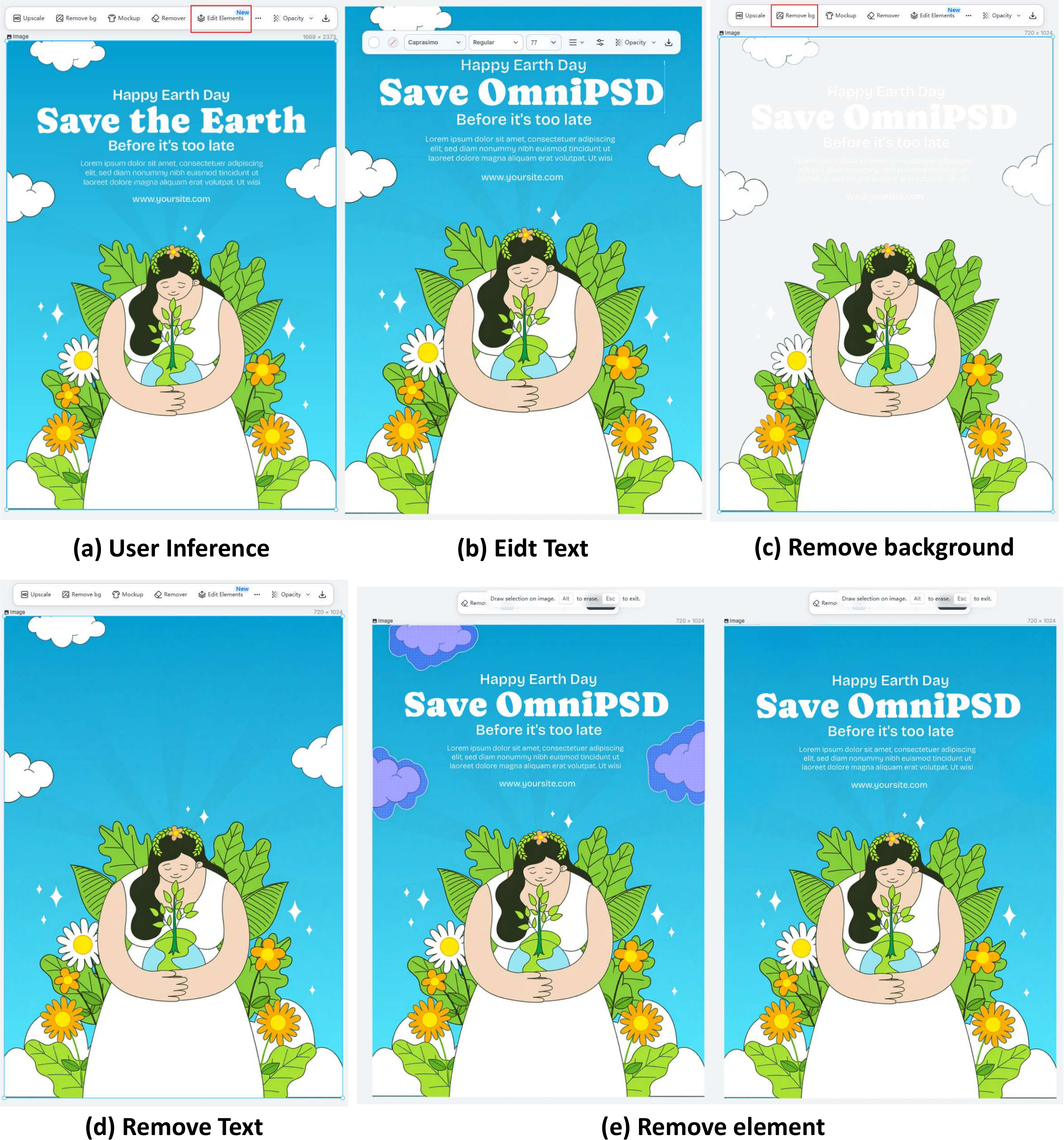}
    \caption{User interface and functional demonstration of OmniPSD. Given a user-uploaded poster image, OmniPSD enables the addition, removal, and editing of textual and graphical elements.}
    \label{fig:ui}
\end{figure*}

%% file: main.bbl
\begin{thebibliography}{80}
\providecommand{\natexlab}[1]{#1}
\providecommand{\url}[1]{\texttt{#1}}
\expandafter\ifx\csname urlstyle\endcsname\relax
  \providecommand{\doi}[1]{doi: #1}\else
  \providecommand{\doi}{doi: \begingroup \urlstyle{rm}\Url}\fi

\bibitem[AI(2024)]{storia_fontclassify}
Storia AI.
\newblock font-classify: Lightweight deep-learning-based font recognition.
\newblock \url{https://github.com/Storia-AI/font-classify}, 2024.
\newblock Accessed: 2025-03-10.

\bibitem[Authors(2023)]{paddleocr}
PaddlePaddle Authors.
\newblock Paddleocr: Open-source ocr toolkit.
\newblock \url{https://github.com/PaddlePaddle/PaddleOCR}, 2023.
\newblock Accessed: 2025-03-10.

\bibitem[Batifol et~al.(2025)Batifol, Blattmann, Boesel, Consul, Diagne, Dockhorn, English, English, Esser, Kulal, et~al.]{batifol2025flux}
Stephen Batifol, Andreas Blattmann, Frederic Boesel, Saksham Consul, Cyril Diagne, Tim Dockhorn, Jack English, Zion English, Patrick Esser, Sumith Kulal, et~al.
\newblock Flux. 1 kontext: Flow matching for in-context image generation and editing in latent space.
\newblock \emph{arXiv e-prints}, pages arXiv--2506, 2025.

\bibitem[Blattmann et~al.(2023)Blattmann, Dockhorn, Kulal, Mendelevitch, Kilian, Lorenz, Levi, English, Voleti, Letts, et~al.]{blattmann2023stable}
Andreas Blattmann, Tim Dockhorn, Sumith Kulal, Daniel Mendelevitch, Maciej Kilian, Dominik Lorenz, Yam Levi, Zion English, Vikram Voleti, Adam Letts, et~al.
\newblock Stable video diffusion: Scaling latent video diffusion models to large datasets.
\newblock \emph{arXiv preprint arXiv:2311.15127}, 2023.

\bibitem[Burgert et~al.(2024)Burgert, Price, Kuen, Li, and Ryoo]{burgert2024magick}
Ryan~D. Burgert, Brian~L. Price, Jason Kuen, Yijun Li, and Michael~S. Ryoo.
\newblock Magick: A large-scale captioned dataset from matting generated images using chroma keying.
\newblock In \emph{Proceedings of the IEEE/CVF Conference on Computer Vision and Pattern Recognition}, 2024.

\bibitem[Cao et~al.(2023)Cao, Wang, Qi, Shan, Qie, and Zheng]{cao2023masactrl}
Mingdeng Cao, Xintao Wang, Zhongang Qi, Ying Shan, Xiaohu Qie, and Yinqiang Zheng.
\newblock Masactrl: Tuning-free mutual self-attention control for consistent image synthesis and editing.
\newblock In \emph{Proceedings of the IEEE/CVF International Conference on Computer Vision}, 2023.

\bibitem[Chen et~al.(2022)Chen, Liu, Wang, Peng, Hao, Chu, Tang, Wu, Chen, Yu, Du, Dang, Hu, and Yu]{chen2022ppmatting}
Guowei Chen, Yi Liu, Jian Wang, Juncai Peng, Yuying Hao, Lutao Chu, Shiyu Tang, Zewu Wu, Zeyu Chen, Zhiliang Yu, Yuning Du, Qingqing Dang, Xiaoguang Hu, and Dianhai Yu.
\newblock Pp-matting: High-accuracy natural image matting.
\newblock \emph{arXiv preprint arXiv:2204.09433}, 2022.

\bibitem[Chen et~al.(2023)Chen, Yu, Ge, Yao, Xie, Wu, Wang, Kwok, Luo, Lu, and Li]{chen2023pixart}
Junsong Chen, Jincheng Yu, Chongjian Ge, Lewei Yao, Enze Xie, Yue Wu, Zhongdao Wang, James Kwok, Ping Luo, Huchuan Lu, and Zhenguo Li.
\newblock Pixart-$\alpha$: Fast training of diffusion transformer for photorealistic text-to-image synthesis.
\newblock \emph{arXiv preprint arXiv:2310.00426}, 2023.

\bibitem[Chen et~al.(2025)Chen, Chen, and Song]{chen2025transanimate}
Xuewei Chen, Zhimin Chen, and Yiren Song.
\newblock Transanimate: Taming layer diffusion to generate rgba video.
\newblock \emph{arXiv preprint arXiv:2503.17934}, 2025.

\bibitem[Cheng et~al.(2025)Cheng, Zhang, Yang, Nie, Li, Wu, and Shao]{cheng2024graphist}
Yutao Cheng, Zhao Zhang, Maoke Yang, Hui Nie, Chunyuan Li, Xinglong Wu, and Jie Shao.
\newblock Graphic design with large multimodal model.
\newblock In \emph{Proceedings of the AAAI Conference on Artificial Intelligence}, 2025.

\bibitem[Dalva et~al.(2024)Dalva, Li, Liu, Zhao, Zhang, Lin, and Yanardag]{dalva2024layerfusion}
Yusuf Dalva, Yijun Li, Qing Liu, Nanxuan Zhao, Jianming Zhang, Zhe Lin, and Pinar Yanardag.
\newblock Layerfusion: Harmonized multi-layer text-to-image generation with generative priors.
\newblock \emph{arXiv preprint arXiv:2412.04460}, 2024.

\bibitem[DeepMind(2025)]{nano_banana}
Google DeepMind.
\newblock Nano-banana (gemini 2.5 flash image): Google deepmind’s image generation and editing model.
\newblock \url{https://aistudio.google.com/models/gemini-2-5-flash-image}, 2025.
\newblock Accessed: 2025-11-14.

\bibitem[Epstein et~al.(2023)Epstein, Jabri, Poole, Efros, and Holynski]{epstein2023diffusionself}
Dave Epstein, Allan Jabri, Ben Poole, Alexei~A. Efros, and Aleksander Holynski.
\newblock Diffusion self-guidance for controllable image generation.
\newblock \emph{Advances in Neural Information Processing Systems}, 36, 2023.

\bibitem[Esser et~al.(2024)Esser, Kulal, Blattmann, Entezari, M{\"u}ller, Saini, Levi, Lorenz, Sauer, Boesel, Podell, Dockhorn, English, Lacey, Goodwin, Marek, and Rombach]{esser2024sd3}
Patrick Esser, Sumith Kulal, Andreas Blattmann, Rahim Entezari, Jonas M{\"u}ller, Harry Saini, Yam Levi, Dominik Lorenz, Axel Sauer, Frederic Boesel, Dustin Podell, Tim Dockhorn, Zion English, Kyle Lacey, Alex Goodwin, Yannik Marek, and Robin Rombach.
\newblock Scaling rectified flow transformers for high-resolution image synthesis.
\newblock \emph{arXiv preprint arXiv:2403.03206}, 2024.

\bibitem[Fontanella et~al.(2024)Fontanella, Tudosiu, Yang, Zhang, and Parisot]{fontanella2024rgba}
Alessandro Fontanella, Petru-Daniel Tudosiu, Yongxin Yang, Shifeng Zhang, and Sarah Parisot.
\newblock Generating compositional scenes via text-to-image rgba instance generation.
\newblock \emph{Advances in Neural Information Processing Systems}, 2024.

\bibitem[Gong et~al.(2025)Gong, Song, Li, Li, and Zhang]{gong2025relationadapter}
Yan Gong, Yiren Song, Yicheng Li, Chenglin Li, and Yin Zhang.
\newblock Relationadapter: Learning and transferring visual relation with diffusion transformers.
\newblock \emph{arXiv preprint arXiv:2506.02528}, 2025.

\bibitem[Goodfellow et~al.(2020)Goodfellow, Pouget-Abadie, Mirza, Xu, Warde-Farley, Ozair, Courville, and Bengio]{goodfellow2020generative}
Ian Goodfellow, Jean Pouget-Abadie, Mehdi Mirza, Bing Xu, David Warde-Farley, Sherjil Ozair, Aaron Courville, and Yoshua Bengio.
\newblock Generative adversarial networks.
\newblock \emph{Communications of the ACM}, 63\penalty0 (11):\penalty0 139--144, 2020.

\bibitem[Hertz et~al.(2023)Hertz, Mokady, Tenenbaum, Aberman, Pritch, and Cohen-Or]{hertz2022prompt}
Amir Hertz, Ron Mokady, Jay Tenenbaum, Kfir Aberman, Yael Pritch, and Daniel Cohen-Or.
\newblock Prompt-to-prompt image editing with cross-attention control.
\newblock In \emph{Proceedings of the International Conference on Learning Representations}, 2023.

\bibitem[Heusel et~al.(2017)Heusel, Ramsauer, Unterthiner, Nessler, and Hochreiter]{fid}
Martin Heusel, Hubert Ramsauer, Thomas Unterthiner, Bernhard Nessler, and Sepp Hochreiter.
\newblock Gans trained by a two time-scale update rule converge to a local nash equilibrium.
\newblock In \emph{Advances in Neural Information Processing Systems}, 2017.

\bibitem[Ho et~al.(2020)Ho, Jain, and Abbeel]{ho2020denoising}
Jonathan Ho, Ajay Jain, and Pieter Abbeel.
\newblock Denoising diffusion probabilistic models.
\newblock \emph{Advances in neural information processing systems}, 33:\penalty0 6840--6851, 2020.

\bibitem[Ho et~al.(2022)Ho, Salimans, Gritsenko, Chan, Norouzi, and Fleet]{ho2022video}
Jonathan Ho, Tim Salimans, Alexey Gritsenko, William Chan, Mohammad Norouzi, and David~J Fleet.
\newblock Video diffusion models.
\newblock \emph{Advances in neural information processing systems}, 35:\penalty0 8633--8646, 2022.

\bibitem[Hu et~al.(2022)Hu, Shen, Wallis, Allen-Zhu, Li, Wang, Wang, Chen, et~al.]{hu2022lora}
Edward~J Hu, Yelong Shen, Phillip Wallis, Zeyuan Allen-Zhu, Yuanzhi Li, Shean Wang, Lu Wang, Weizhu Chen, et~al.
\newblock Lora: Low-rank adaptation of large language models.
\newblock \emph{ICLR}, 1\penalty0 (2):\penalty0 3, 2022.

\bibitem[Hu et~al.(2024)Hu, Peng, Luo, Ji, Peng, Jiang, Zhang, Jin, Wang, and Ji]{hu2024diffumatting}
Xiaobin Hu, Xu Peng, Donghao Luo, Xiaozhong Ji, Jinlong Peng, Zhengkai Jiang, Jiangning Zhang, Taisong Jin, Chengjie Wang, and Rongrong Ji.
\newblock Diffumatting: Synthesizing arbitrary objects with matting-level annotation.
\newblock \emph{arXiv preprint arXiv:2403.06168}, 2024.

\bibitem[Huang et~al.(2025{\natexlab{a}})Huang, Li, Zhao, Pan, Zeng, and Dai]{psdiffusion2025}
Dingbang Huang, Wenbo Li, Yifei Zhao, Xinyu Pan, Yanhong Zeng, and Bo Dai.
\newblock Psdiffusion: Harmonized multi-layer image generation via layout and appearance alignment.
\newblock \emph{arXiv preprint arXiv:2505.11468}, 2025{\natexlab{a}}.

\bibitem[Huang et~al.(2024)Huang, Cai, Han, Liang, Zhang, Xu, and Xu]{yang2023layerdiff}
Runhui Huang, Kaixin Cai, Jianhua Han, Xiaodan Liang, Wei Zhang, Songcen Xu, and Hang Xu.
\newblock Layerdiff: Exploring text-guided multi-layered composable image synthesis via layer-collaborative diffusion model.
\newblock \emph{arXiv preprint arXiv:2403.12036}, 2024.

\bibitem[Huang et~al.(2025{\natexlab{b}})Huang, Song, Zhang, Guo, Wang, and Liu]{huang2025arteditor}
Shijie Huang, Yiren Song, Yuxuan Zhang, Hailong Guo, Xueyin Wang, and Jiaming Liu.
\newblock Arteditor: Learning customized instructional image editor from few-shot examples.
\newblock In \emph{Proceedings of the IEEE/CVF International Conference on Computer Vision}, pages 17651--17662, 2025{\natexlab{b}}.

\bibitem[Inoue et~al.(2024)Inoue, Masui, Shimoda, and Yamaguchi]{inoue2024opencole}
Naoto Inoue, Kento Masui, Wataru Shimoda, and Kota Yamaguchi.
\newblock Opencole: Towards reproducible automatic graphic design generation.
\newblock \emph{arXiv preprint arXiv:2406.08232}, 2024.

\bibitem[Jia et~al.(2023)Jia, Li, Liu, Shen, Chen, Yuan, Zheng, Chen, Li, Xie, Zhang, and Guo]{jia2023cole}
Peidong Jia, Chenxuan Li, Zeyu Liu, Yichao Shen, Xingru Chen, Yuhui Yuan, Yinglin Zheng, Dong Chen, Ji Li, Xiaodong Xie, Shanghang Zhang, and Baining Guo.
\newblock Cole: A hierarchical generation framework for graphic design.
\newblock \emph{arXiv preprint arXiv:2311.16974}, 2023.

\bibitem[Jiang et~al.(2025)Jiang, Gu, Song, Tsang, and Shou]{jiang2025personalized}
Yuxin Jiang, Yuchao Gu, Yiren Song, Ivor Tsang, and Mike~Zheng Shou.
\newblock Personalized vision via visual in-context learning.
\newblock \emph{arXiv preprint arXiv:2509.25172}, 2025.

\bibitem[Kang et~al.(2025)Kang, Sim, Kim, Kim, Nam, and Cho]{kang2025layeringdiff}
Kyoungkook Kang, Gyujin Sim, Geonung Kim, Donguk Kim, Seungho Nam, and Sunghyun Cho.
\newblock Layeringdiff: Layered image synthesis via generation, then disassembly with generative knowledge.
\newblock \emph{arXiv preprint arXiv:2501.01197}, 2025.

\bibitem[Kikuchi et~al.(2025)Kikuchi, Honda, Inoue, Otani, Simo-Serra, and Yamaguchi]{kikuchi2025markupdm}
Kotaro Kikuchi, Ukyo Honda, Naoto Inoue, Mayu Otani, Edgar Simo-Serra, and Kota Yamaguchi.
\newblock Multimodal markup document models for graphic design completion.
\newblock In \emph{Proceedings of the ACM International Conference on Multimedia}, 2025.

\bibitem[Lee et~al.(2020)Lee, Chang, Peng, and Hang]{lee2020hybrid}
Wei-Cheng Lee, Chih-Peng Chang, Wen-Hsiao Peng, and Hsueh-Ming Hang.
\newblock A hybrid layered image compressor with deep-learning technique.
\newblock In \emph{IEEE International Workshop on Multimedia Signal Processing (MMSP)}, 2020.

\bibitem[Li et~al.(2022)Li, Zhang, Maybank, and Tao]{li2022bridging}
Jizhizi Li, Jing Zhang, Stephen~J. Maybank, and Dacheng Tao.
\newblock Bridging composite and real: Towards end-to-end deep image matting.
\newblock \emph{International Journal of Computer Vision}, 130\penalty0 (2):\penalty0 246--266, 2022.

\bibitem[Li et~al.(2024{\natexlab{a}})Li, Jain, and Shi]{li2024mattinganything}
Jiachen Li, Jitesh Jain, and Humphrey Shi.
\newblock Matting anything.
\newblock In \emph{Proceedings of the IEEE/CVF Conference on Computer Vision and Pattern Recognition Workshops}, pages 1775--1785, 2024{\natexlab{a}}.

\bibitem[Li et~al.(2023)Li, Liu, Wu, Mu, Yang, Gao, Li, and Lee]{li2023gligen}
Yuheng Li, Haotian Liu, Qingyang Wu, Fangzhou Mu, Jianwei Yang, Jianfeng Gao, Chunyuan Li, and Yong~Jae Lee.
\newblock Gligen: Open-set grounded text-to-image generation.
\newblock In \emph{Proceedings of the IEEE/CVF Conference on Computer Vision and Pattern Recognition}, 2023.

\bibitem[Li et~al.(2024{\natexlab{b}})Li, Zhang, Lin, Xiong, Long, Deng, Zhang, Liu, Huang, Xiao, Chen, He, Li, Li, Zhang, Quan, Lu, Huang, Yuan, Zheng, Li, Zhang, Zhang, Chen, Liu, Fang, Wang, Xue, Tao, Zhu, Liu, Lin, Sun, Li, Wang, Chen, Hu, Xiao, Chen, Liu, Liu, Wang, Yang, Jiang, and Lu]{hunyuan2024}
Zhimin Li, Jianwei Zhang, Qin Lin, Jiangfeng Xiong, Yanxin Long, Xinchi Deng, Yingfang Zhang, Xingchao Liu, Minbin Huang, Zedong Xiao, Dayou Chen, Jiajun He, Jiahao Li, Wenyue Li, Chen Zhang, Rongwei Quan, Jianxiang Lu, Jiabin Huang, Xiaoyan Yuan, Xiaoxiao Zheng, Yixuan Li, Jihong Zhang, Chao Zhang, Meng Chen, Jie Liu, Zheng Fang, Weiyan Wang, Jinbao Xue, Yangyu Tao, Jianchen Zhu, Kai Liu, Sihuan Lin, Yifu Sun, Yun Li, Dongdong Wang, Mingtao Chen, Zhichao Hu, Xiao Xiao, Yan Chen, Yuhong Liu, Wei Liu, Di Wang, Yong Yang, Jie Jiang, and Qinglin Lu.
\newblock Hunyuan-dit: A powerful multi-resolution diffusion transformer with fine-grained chinese understanding.
\newblock \emph{arXiv preprint arXiv:2405.08748}, 2024{\natexlab{b}}.

\bibitem[Lipman et~al.(2022)Lipman, Chen, Ben-Hamu, Nickel, and Le]{lipman2022flow}
Yaron Lipman, Ricky~TQ Chen, Heli Ben-Hamu, Maximilian Nickel, and Matt Le.
\newblock Flow matching for generative modeling.
\newblock \emph{arXiv preprint arXiv:2210.02747}, 2022.

\bibitem[Liu et~al.(2022)Liu, Gong, and Liu]{liu2022flow}
Xingchao Liu, Chengyue Gong, and Qiang Liu.
\newblock Flow straight and fast: Learning to generate and transfer data with rectified flow.
\newblock \emph{arXiv preprint arXiv:2209.03003}, 2022.

\bibitem[Liu et~al.(2020)Liu, Lai, Yang, Chuang, and Huang]{liu2022learning}
Yu-Lun Liu, Wei-Sheng Lai, Ming-Hsuan Yang, Yung-Yu Chuang, and Jia-Bin Huang.
\newblock Learning to see through obstructions with layered decomposition.
\newblock In \emph{Proceedings of the IEEE/CVF Conference on Computer Vision and Pattern Recognition (CVPR)}, 2020.

\bibitem[Ma et~al.(2024{\natexlab{a}})Ma, He, Cun, Wang, Chen, Li, and Chen]{ma2024followpose}
Yue Ma, Yingqing He, Xiaodong Cun, Xintao Wang, Siran Chen, Xiu Li, and Qifeng Chen.
\newblock Follow your pose: Pose-guided text-to-video generation using pose-free videos.
\newblock In \emph{Proceedings of the AAAI Conference on Artificial Intelligence}, pages 4117--4125, 2024{\natexlab{a}}.

\bibitem[Ma et~al.(2024{\natexlab{b}})Ma, Liu, Wang, Pan, He, Yuan, Zeng, Cai, Shum, Liu, et~al.]{ma2024followyouremoji}
Yue Ma, Hongyu Liu, Hongfa Wang, Heng Pan, Yingqing He, Junkun Yuan, Ailing Zeng, Chengfei Cai, Heung-Yeung Shum, Wei Liu, et~al.
\newblock Follow-your-emoji: Fine-controllable and expressive freestyle portrait animation.
\newblock In \emph{SIGGRAPH Asia 2024 Conference Papers}, pages 1--12, 2024{\natexlab{b}}.

\bibitem[Ma et~al.(2025{\natexlab{a}})Ma, He, Wang, Wang, Shen, Qi, Ying, Cai, Li, Shum, et~al.]{ma2025followyourclick}
Yue Ma, Yingqing He, Hongfa Wang, Andong Wang, Leqi Shen, Chenyang Qi, Jixuan Ying, Chengfei Cai, Zhifeng Li, Heung-Yeung Shum, et~al.
\newblock Follow-your-click: Open-domain regional image animation via motion prompts.
\newblock In \emph{Proceedings of the AAAI Conference on Artificial Intelligence}, pages 6018--6026, 2025{\natexlab{a}}.

\bibitem[Ma et~al.(2025{\natexlab{b}})Ma, Liu, Zhu, Yang, Feng, Zhang, Li, Han, Qi, and Chen]{ma2025followyourmotion}
Yue Ma, Yulong Liu, Qiyuan Zhu, Ayden Yang, Kunyu Feng, Xinhua Zhang, Zhifeng Li, Sirui Han, Chenyang Qi, and Qifeng Chen.
\newblock Follow-your-motion: Video motion transfer via efficient spatial-temporal decoupled finetuning.
\newblock \emph{arXiv preprint arXiv:2506.05207}, 2025{\natexlab{b}}.

\bibitem[OpenAI(2023)]{gpt}
OpenAI.
\newblock Gpt-4 technical report.
\newblock \emph{arXiv preprint arXiv:2303.08774}, 2023.

\bibitem[Pan et~al.(2020)Pan, Luo, Yang, and Li]{mma}
Zexu Pan, Zhaojie Luo, Jichen Yang, and Haizhou Li.
\newblock Multi-modal attention for speech emotion recognition.
\newblock \emph{arXiv preprint arXiv:2009.04107}, 2020.

\bibitem[Peebles and Xie(2023)]{peebles2023scalable}
William Peebles and Saining Xie.
\newblock Scalable diffusion models with transformers.
\newblock In \emph{Proceedings of the IEEE/CVF international conference on computer vision}, pages 4195--4205, 2023.

\bibitem[Podell et~al.(2023)Podell, English, Lacey, Blattmann, Dockhorn, M{\"u}ller, Penna, and Rombach]{podell2023sdxl}
Dustin Podell, Zion English, Kyle Lacey, Andreas Blattmann, Tim Dockhorn, Jonas M{\"u}ller, Joe Penna, and Robin Rombach.
\newblock Sdxl: Improving latent diffusion models for high-resolution image synthesis.
\newblock \emph{arXiv preprint arXiv:2307.01952}, 2023.

\bibitem[Pu et~al.(2025)Pu, Zhao, Tang, Yin, Ye, Yuan, Chen, Bao, Zhang, Wang, et~al.]{pu2025art}
Yifan Pu, Yiming Zhao, Zhicong Tang, Ruihong Yin, Haoxing Ye, Yuhui Yuan, Dong Chen, Jianmin Bao, Sirui Zhang, Yanbin Wang, et~al.
\newblock Art: Anonymous region transformer for variable multi-layer transparent image generation.
\newblock In \emph{Proceedings of the Computer Vision and Pattern Recognition Conference}, pages 7952--7962, 2025.

\bibitem[Quattrini et~al.(2024)Quattrini, Pippi, Cascianelli, and Cucchiara]{quattrini2024alfie}
Fabio Quattrini, Vittorio Pippi, Silvia Cascianelli, and Rita Cucchiara.
\newblock Alfie: Democratising rgba image generation with no \$\$\$.
\newblock \emph{arXiv preprint arXiv:2408.14826}, 2024.

\bibitem[Radford et~al.(2021)Radford, Kim, Hallacy, Ramesh, Goh, Agarwal, Sastry, Askell, Mishkin, et~al.]{clip}
Alec Radford, Jong~Wook Kim, Chris Hallacy, Aditya Ramesh, Gabriel Goh, Sandhini Agarwal, Girish Sastry, Amanda Askell, Pamela Mishkin, et~al.
\newblock Learning transferable visual models from natural language supervision.
\newblock In \emph{International Conference on Machine Learning}, 2021.

\bibitem[Ravi et~al.(2024)Ravi, Gabeur, Hu, Hu, Ryali, Ma, Khedr, R{\"a}dle, Rolland, Gustafson, Mintun, Pan, Alwala, Carion, Wu, Girshick, Doll{\'a}r, and Feichtenhofer]{ravi2024sam2}
Nikhila Ravi, Valentin Gabeur, Yuan-Ting Hu, Ronghang Hu, Chaitanya Ryali, Tengyu Ma, Haitham Khedr, Roman R{\"a}dle, Chloe Rolland, Laura Gustafson, Eric Mintun, Junting Pan, Kalyan~Vasudev Alwala, Nicolas Carion, Chao-Yuan Wu, Ross Girshick, Piotr Doll{\'a}r, and Christoph Feichtenhofer.
\newblock Sam 2: Segment anything in images and videos.
\newblock \emph{arXiv preprint arXiv:2408.00714}, 2024.

\bibitem[Rombach et~al.(2022)Rombach, Blattmann, Lorenz, Esser, and Ommer]{rombach2022high}
Robin Rombach, Andreas Blattmann, Dominik Lorenz, Patrick Esser, and Bj{\"o}rn Ommer.
\newblock High-resolution image synthesis with latent diffusion models.
\newblock In \emph{Proceedings of the IEEE/CVF conference on computer vision and pattern recognition}, pages 10684--10695, 2022.

\bibitem[Sengupta et~al.(2020)Sengupta, Jayaram, Curless, Seitz, and Kemelmacher-Shlizerman]{sengupta2020background}
Soumyadip Sengupta, Vivek Jayaram, Brian Curless, Steven~M Seitz, and Ira Kemelmacher-Shlizerman.
\newblock Background matting: The world is your green screen.
\newblock In \emph{Proceedings of the IEEE/CVF Conference on Computer Vision and Pattern Recognition}, pages 2291--2300, 2020.

\bibitem[Shabani et~al.(2024)Shabani, Wang, Liu, Zhao, Yang, and Furukawa]{shabani2024vlc}
Mohammad~Amin Shabani, Zhaowen Wang, Difan Liu, Nanxuan Zhao, Jimei Yang, and Yasutaka Furukawa.
\newblock Visual layout composer: Image-vector dual diffusion model for design layout generation.
\newblock In \emph{Proceedings of the IEEE/CVF Conference on Computer Vision and Pattern Recognition}, 2024.

\bibitem[Song et~al.(2020)Song, Meng, and Ermon]{song2020denoising}
Jiaming Song, Chenlin Meng, and Stefano Ermon.
\newblock Denoising diffusion implicit models.
\newblock \emph{arXiv preprint arXiv:2010.02502}, 2020.

\bibitem[Song et~al.(2024)Song, Huang, Yao, Ye, Ci, Liu, Zhang, and Shou]{song2024processpainter}
Yiren Song, Shijie Huang, Chen Yao, Xiaojun Ye, Hai Ci, Jiaming Liu, Yuxuan Zhang, and Mike~Zheng Shou.
\newblock Processpainter: Learn painting process from sequence data.
\newblock \emph{arXiv preprint arXiv:2406.06062}, 2024.

\bibitem[Song et~al.(2025{\natexlab{a}})Song, Chen, and Shou]{song2025layertracer}
Yiren Song, Danze Chen, and Mike~Zheng Shou.
\newblock Layertracer: Cognitive-aligned layered svg synthesis via diffusion transformer.
\newblock \emph{arXiv preprint arXiv:2502.01105}, 2025{\natexlab{a}}.

\bibitem[Song et~al.(2025{\natexlab{b}})Song, Liu, and Shou]{song2025makeanything}
Yiren Song, Cheng Liu, and Mike~Zheng Shou.
\newblock Makeanything: Harnessing diffusion transformers for multi-domain procedural sequence generation.
\newblock \emph{arXiv preprint arXiv:2502.01572}, 2025{\natexlab{b}}.

\bibitem[Song et~al.(2025{\natexlab{c}})Song, Liu, and Shou]{song2025omniconsistency}
Yiren Song, Cheng Liu, and Mike~Zheng Shou.
\newblock Omniconsistency: Learning style-agnostic consistency from paired stylization data.
\newblock \emph{arXiv preprint arXiv:2505.18445}, 2025{\natexlab{c}}.

\bibitem[Tudosiu et~al.(2024)Tudosiu, Yang, Zhang, Chen, McDonagh, Lampouras, Iacobacci, and Parisot]{tudosiu2024mulan}
Petru-Daniel Tudosiu, Yongxin Yang, Shifeng Zhang, Fei Chen, Steven McDonagh, Gerasimos Lampouras, Ignacio Iacobacci, and Sarah Parisot.
\newblock Mulan: A multi layer annotated dataset for controllable text-to-image generation.
\newblock In \emph{Proceedings of the IEEE/CVF Conference on Computer Vision and Pattern Recognition}, 2024.

\bibitem[Wan et~al.(2024)Wan, Luo, Cai, Song, Zhao, Bai, He, and Gong]{wan2024grid}
Cong Wan, Xiangyang Luo, Zijian Cai, Yiren Song, Yunlong Zhao, Yifan Bai, Yuhang He, and Yihong Gong.
\newblock Grid: Visual layout generation.
\newblock \emph{arXiv preprint arXiv:2412.10718}, 2024.

\bibitem[Wang et~al.(2024)Wang, Darrell, Rambhatla, Girdhar, and Misra]{wang2024instancediffusion}
Xudong Wang, Trevor Darrell, Sai~Saketh Rambhatla, Rohit Girdhar, and Ishan Misra.
\newblock Instancediffusion: Instance-level control for image generation.
\newblock In \emph{Proceedings of the IEEE/CVF Conference on Computer Vision and Pattern Recognition}, 2024.

\bibitem[Wang et~al.(2025{\natexlab{a}})Wang, Fu, Huang, He, and Jiang]{wang2025msdiffusion}
Xierui Wang, Siming Fu, Qihan Huang, Wanggui He, and Hao Jiang.
\newblock Ms-diffusion: Multi-subject zero-shot image personalization with layout guidance.
\newblock In \emph{Proceedings of the International Conference on Learning Representations}, 2025{\natexlab{a}}.

\bibitem[Wang et~al.(2025{\natexlab{b}})Wang, Yu, Zhan, and Yuan]{alphavae2025}
Zile Wang, Hao Yu, Jiabo Zhan, and Chun Yuan.
\newblock Alphavae: Unified end-to-end rgba image reconstruction and generation with alpha-aware representation learning.
\newblock \emph{arXiv preprint arXiv:2507.09308}, 2025{\natexlab{b}}.

\bibitem[Wang et~al.(2025{\natexlab{c}})Wang, Zhao, Zhou, Lu, Li, and Song]{wang2025diffdecompose}
Zitong Wang, Hang Zhao, Qianyu Zhou, Xuequan Lu, Xiangtai Li, and Yiren Song.
\newblock Diffdecompose: Layer-wise decomposition of alpha-composited images via diffusion transformers.
\newblock \emph{arXiv preprint arXiv:2505.21541}, 2025{\natexlab{c}}.

\bibitem[Weng et~al.(2024)Weng, Huang, Qiao, Hu, Lin, Zhang, and Chen]{weng2024desigen}
Haohan Weng, Danqing Huang, Yu Qiao, Zheng Hu, Chin-Yew Lin, Tong Zhang, and C.~L.~Philip Chen.
\newblock Desigen: A pipeline for controllable design template generation.
\newblock In \emph{Proceedings of the IEEE/CVF Conference on Computer Vision and Pattern Recognition}, 2024.

\bibitem[Xiang and Sun(2025)]{layerdiffuseflux2025}
Qiang Xiang and Shuang Sun.
\newblock Layerdiffuse-flux: Flux version implementation of layerdiffusion.
\newblock \url{https://github.com/FireRedTeam/LayerDiffuse-Flux}, 2025.
\newblock Code repository, accessed 2025-11-13.

\bibitem[Xu et~al.(2017)Xu, Price, Cohen, and Huang]{xu2017deep}
Ning Xu, Brian Price, Scott Cohen, and Thomas Huang.
\newblock Deep image matting.
\newblock In \emph{Proceedings of the IEEE conference on computer vision and pattern recognition}, pages 2970--2979, 2017.

\bibitem[Yang et~al.(2024{\natexlab{a}})Yang, Liu, Deng, Kim, Mei, Shen, and Chen]{flux2024}
Chenglin Yang, Celong Liu, Xueqing Deng, Dongwon Kim, Xing Mei, Xiaohui Shen, and Liang-Chieh Chen.
\newblock 1.58-bit flux.
\newblock \emph{arXiv preprint arXiv:2412.18653}, 2024{\natexlab{a}}.

\bibitem[Yang et~al.(2024{\natexlab{b}})Yang, Liu, Li, Kim, Pakhomov, Ren, Zhang, Lin, Xie, and Zhou]{yang2024layerdecomp}
Jinrui Yang, Qing Liu, Yijun Li, Soo~Ye Kim, Daniil Pakhomov, Mengwei Ren, Jianming Zhang, Zhe Lin, Cihang Xie, and Yuyin Zhou.
\newblock Generative image layer decomposition with visual effects.
\newblock \emph{arXiv preprint arXiv:2411.17864}, 2024{\natexlab{b}}.

\bibitem[Yang et~al.(2024{\natexlab{c}})Yang, Luo, Qi, Wu, Shan, and Chen]{yang2024posterllava}
Tao Yang, Yingmin Luo, Zhongang Qi, Yang Wu, Ying Shan, and Chang~Wen Chen.
\newblock Posterllava: Constructing a unified multi-modal layout generator with llm.
\newblock \emph{arXiv preprint arXiv:2406.02884}, 2024{\natexlab{c}}.

\bibitem[Ye et~al.(2023)Ye, Zhang, Liu, Han, and Yang]{ye2023ip}
Hu Ye, Jun Zhang, Sibo Liu, Xiao Han, and Wei Yang.
\newblock Ip-adapter: Text compatible image prompt adapter for text-to-image diffusion models.
\newblock \emph{arXiv preprint arXiv:2308.06721}, 2023.

\bibitem[Yu et~al.(2023)Yu, Feng, Feng, Liu, Jin, Zeng, and Chen]{yu2023inpaintanything}
Tao Yu, Runseng Feng, Ruoyu Feng, Jinming Liu, Xin Jin, Wenjun Zeng, and Zhibo Chen.
\newblock Inpaint anything: Segment anything meets image inpainting.
\newblock \emph{arXiv preprint arXiv:2304.06790}, 2023.

\bibitem[Zhang and Zhang(2024{\natexlab{a}})]{layerdiffuse2023}
Lvmin Zhang and Richard Zhang.
\newblock Transparent image layer diffusion using latent transparency.
\newblock \emph{arXiv preprint arXiv:2402.17113}, 2024{\natexlab{a}}.

\bibitem[Zhang and Zhang(2024{\natexlab{b}})]{zhang2024layerdiffuse}
Lvmin Zhang and Richard Zhang.
\newblock Transparent image layer diffusion using latent transparency.
\newblock \emph{arXiv preprint arXiv:2402.17113}, 2024{\natexlab{b}}.

\bibitem[Zhang et~al.(2023{\natexlab{a}})Zhang, Rao, and Agrawala]{zhang2023controlnet}
Lvmin Zhang, Anyi Rao, and Maneesh Agrawala.
\newblock Adding conditional control to text-to-image diffusion models.
\newblock In \emph{Proceedings of the IEEE/CVF International Conference on Computer Vision}, pages 3836--3847, 2023{\natexlab{a}}.

\bibitem[Zhang et~al.(2023{\natexlab{b}})Zhang, Zhao, Lu, and Chien]{zhang2023text2layer}
Xinyang Zhang, Wentian Zhao, Xin Lu, and Jeff Chien.
\newblock Text2layer: Layered image generation using latent diffusion model.
\newblock \emph{arXiv preprint arXiv:2307.09781}, 2023{\natexlab{b}}.

\bibitem[Zhang et~al.(2024{\natexlab{a}})Zhang, Song, Liu, Wang, Yu, Tang, Li, Tang, Hu, Pan, et~al.]{zhang2024ssr}
Yuxuan Zhang, Yiren Song, Jiaming Liu, Rui Wang, Jinpeng Yu, Hao Tang, Huaxia Li, Xu Tang, Yao Hu, Han Pan, et~al.
\newblock Ssr-encoder: Encoding selective subject representation for subject-driven generation.
\newblock In \emph{Proceedings of the IEEE/CVF Conference on Computer Vision and Pattern Recognition}, pages 8069--8078, 2024{\natexlab{a}}.

\bibitem[Zhang et~al.(2024{\natexlab{b}})Zhang, Wei, Zhang, Song, Liu, Li, Tang, Hu, and Zhao]{zhang2024stable}
Yuxuan Zhang, Lifu Wei, Qing Zhang, Yiren Song, Jiaming Liu, Huaxia Li, Xu Tang, Yao Hu, and Haibo Zhao.
\newblock Stable-makeup: When real-world makeup transfer meets diffusion model.
\newblock \emph{arXiv preprint arXiv:2403.07764}, 2024{\natexlab{b}}.

\bibitem[Zhang et~al.(2025)Zhang, Zhang, Song, Zhang, Tang, and Liu]{zhang2025stable}
Yuxuan Zhang, Qing Zhang, Yiren Song, Jichao Zhang, Hao Tang, and Jiaming Liu.
\newblock Stable-hair: Real-world hair transfer via diffusion model.
\newblock In \emph{Proceedings of the AAAI Conference on Artificial Intelligence}, pages 10348--10356, 2025.

\end{thebibliography}
